\documentclass[journal]{IEEEtran}
\IEEEoverridecommandlockouts
\usepackage[utf8]{inputenc}
\usepackage[T1]{fontenc}
\usepackage{graphicx}
\usepackage{booktabs}
\usepackage{color}

\usepackage[cmex10]{amsmath}
\usepackage{multirow}
\usepackage{array}
\usepackage{subfig}
\newcommand\redbf[1]{\textcolor{red}{\textbf{#1}}}

\graphicspath{ {./ImagesJPG2/} }

\begin{document}

\title{Spatial and Angular Resolution Enhancement of Light Fields Using Convolutional Neural Networks}

\author{M. Shahzeb Khan Gul
        and Bahadir K. Gunturk \\
        Dept. of Electrical and Electronics Engineering, Istanbul Medipol University, Istanbul, Turkey \\
        mskhangul@st.medipol.edu.tr, bkgunturk@medipol.edu.tr
}

\maketitle

\begin{abstract}
Light field imaging extends the traditional photography by capturing both spatial and angular distribution of light, which enables new capabilities, including post-capture refocusing, post-capture aperture control, and depth estimation from a single shot. Micro-lens array (MLA) based light field cameras offer a cost-effective approach to capture light field. A major drawback of MLA based light field cameras is low spatial resolution, which is due to the fact that a single image sensor is shared to capture both spatial and angular information. In this paper, we present a learning based light field enhancement approach. Both spatial and angular resolution of captured light field is enhanced using convolutional neural networks. The proposed method is tested with real light field data captured with a Lytro light field camera, clearly demonstrating spatial and angular resolution improvement.     

\end{abstract}



\section{Introduction}
Light field refers to the collection of light rays in 3D space. With a light field imaging system, light rays in different directions are recorded separately, unlike a traditional imaging system, where a pixel records the total amount of light received by the lens regardless of the direction. The angular information enables new capabilities, including depth estimation, post-capture refocusing, post-capture aperture size and shape control, and 3D modelling. Light field imaging can be used in different application areas, including 3D optical inspection, robotics, microscopy, photography, and computer graphics.    

Light field imaging is first described by Lippmann, who proposed to use a set of small biconvex lenses to capture light rays in different directions and refers to it as integral imaging \cite{lippmann1908}. The term "light field" was first used by Gershun, who studied the radiometric properties of light in space \cite{gershun1936light}. Adelson and Bergen used the term "plenoptic function" and defined it as the function of light rays in terms of intensity, position in space, travel direction, wavelength, and time \cite{adelson1991plenoptic}. Adelson and Wang described and implemented a light field camera that incorporates a single main lens along with a micro-lens array \cite{adelson1992}. This design approach is later adopted in commercial light field cameras \cite{Lytro,Raytrix}. In 1996, Levoy and Hanrahan \cite{levoy1996light} and Gortler \textit{et al.} \cite{gortler1996}  formulated light field as a 4D function, and studied ray space representation and light field re-sampling. Over the years, light field imaging theory and applications have continued to be developed further. Key developments include post-capture refocusing \cite{isaksen2000}, Fourier-domain light field processing \cite{ng2005fourier}, light field microscopy \cite{levoy2006}, focused plenoptic camera \cite{lumsdaine2009focused}, and multi-focus plenoptic camera \cite{perwass2012}. 

\begin{figure}
\centering
   \includegraphics[width=0.39\textwidth]{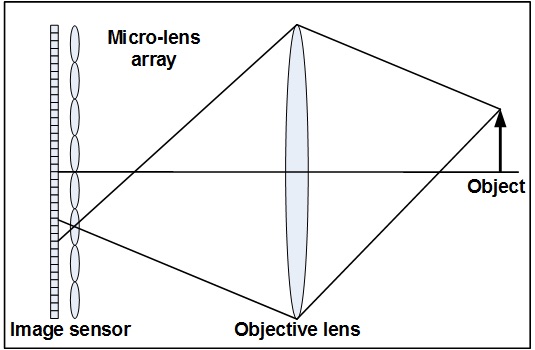}\\
   \includegraphics[width=0.385\textwidth]{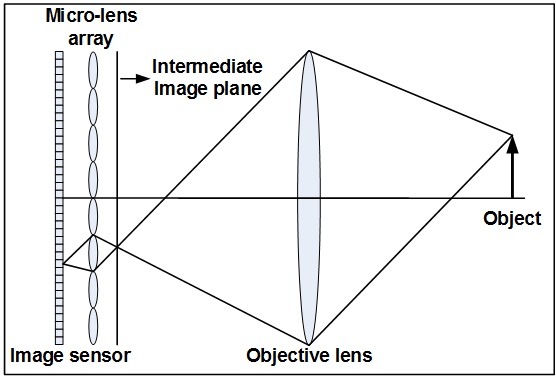}
      \caption{Two main approaches for MLA-based light field camera design. Top: The distance between the image sensor and the MLA is equal to the focal length of a micro-lens (lenslet) in the MLA. Bottom: The objective lens forms an image of the scene on an intermediate image plane, which is then relayed by the lenslets to the image sensor.}
      \label{structureLytro}
\end{figure}

Light field acquisition can be done in various ways, such as camera arrays \cite{wilburn2005high}, optical masks \cite{veeraraghavan2007dappled}, angle-sensitive pixels \cite{wang2011angle}, and micro-lens arrays \cite{ng2005fourier,lumsdaine2009focused}. Among these different approaches, micro-lens array (MLA) based light field cameras provide a cost-effective solution, and have been successfully commercialized \cite{Lytro,Raytrix}. There are two basic implementation approaches of MLA-based light field cameras. In one approach, the image sensor is placed at the focal length of the micro-lenses \cite{ng2005fourier,Lytro}. In the other approach, a micro-lens relays the image (formed by the objective lens on an intermediate image plane) to the image sensor \cite{lumsdaine2009focused,Raytrix}. These two approaches are illustrated in Figure \ref{structureLytro}. In the first approach, the sensor pixels behind a micro-lens (also called a lenslet) on the MLA record light rays coming from different directions. Each lenslet region provides a single pixel value for a perspective image; therefore, the number of lenslets corresponds to the number of pixels in a perspective image. That is, the spatial resolution is defined by the number of lenslets in the MLA. The number of pixels behind a lenslet, on the other hand, defines the angular resolution, that is, the number of perspective images. In the second approach, a lenslet forms an image of the scene from a particular viewpoint. The number of lenslets defines the angular resolution; and, the number of pixels behind a lenslet gives the spatial resolution of a perspective image.   

In the MLA-based light field cameras, there is a trade-off between spatial resolution and angular resolution, since a single image sensor is used to capture both. For example, in the first generation Lytro camera, an 11 megapixel image sensor produces 11x11 sub-aperture perspective images, each with a spatial resolution of about 0.1 megapixels. Such a low spatial resolution prevents the widespread adoption of light field cameras. In recent years, different methods have been proposed to tackle the low spatial resolution issue. Hybrid systems, consisting of a light field sensor and a regular sensor, have been presented \cite{boominathan2014improving,wang2016high,alam2016hybrid}, where the high spatial resolution image from the regular sensor is used to enhance the light field sub-aperture (perspective) images. The disadvantages of hybrid systems include increased cost and larger camera dimensions. Another approach is to apply multi-frame super-resolution techniques to the sub-aperture images of a light field \cite{bishop2009light,wanner2012spatial}. It is also possible to apply learning-based super-resolution techniques to each sub-aperture image of a light field \cite{cho2013modeling}.    

In this paper, we present a convolutional neural network based light field super-resolution method. The method has two sub-networks; one is trained to increase the angular resolution, that is, to synthesize novel viewpoints (sub-aperture images); and the other is trained to increase the spatial resolution of each sub-aperture image. We show that the proposed method provides significant increase in image quality, visually as well as quantitatively (in terms of peak signal-to-noise ratio and structural similarity index \cite{wang2004image}), and improves depth estimation accuracy. 

The paper is organized as follows. We present the related work in the literature in Section II. We explain the proposed method in Section III, present our experimental results in Section IV, and conclude the paper in Section V.

\section{Related Work}
\subsection{Super-Resolution of Light Field}
One approach to enhance the spatial resolution of images captured with an MLA-based light field camera is to apply a multi-frame super-resolution technique on the perspective images obtained from the light field capture. The Bayesian super-resolution restoration framework is commonly used, with Lambertian and textual priors \cite{bishop2009light}, Gaussian mixture models \cite{mitra2012light}, and variational models \cite{wanner2012spatial}.

Learning-based single-image super-resolution methods can also be adopted to address the low spatial resolution issue of light fields. In \cite{cho2013modeling}, a dictionary learning based super-resolution method is presented, demonstrating a clear improvement over standard interpolation techniques when converting raw light field capture into perspective images. Another learning based method is presented in \cite{yoon2015learning}, which incorporates deep convolutional neural networks for spatial and angular resolution enhancement of light fields. Alternative to spatial domain resolution enhancement approaches, frequency domain methods, utilizing signal sparsity and Fourier slice theorem, have also been proposed \cite{perez2012fourier,shi2014light}. 

In contrast to single-sensor light field imaging systems, hybrid light field imaging system have also been introduced to improve spatial resolution. In the hybrid imaging system proposed by Boominathan \textit{et al.} \cite{boominathan2014improving}, a patch-based algorithm is used to super-resolve low-resolution light field views using high-resolution patches acquired from a standard high-resolution camera. There are several other hybrid imaging system presented \cite{wang2016high,wu2015novel,alam2016hybrid}, combining images from a standard camera and a light field camera. Among these, the work in \cite{alam2016hybrid} demonstrates a wide baseline hybrid stereo system, improving range and accuracy of depth estimation in addition to spatial resolution enhancement.

\subsection{Deep Learning for Image Restoration}
Convolutional neural networks (CNNs) are variants of multi-layer perceptron networks. Convolution layer, which is inspired from the work of Hubel and Wiesel \cite{hubel1968receptive} showing that visual neurons respond to local regions, is the fundamental part of a CNN. In \cite{lecun1998gradient}, LeCun \textit{et al.} presented a convolutional neural network based pattern recognition algorithm, promoting further research in this field. Deep learning with convolutional neural networks has been extensively and successfully applied to computer vision applications. While most of these applications are on classification and object recognition, there are also deep-learning based low-level vision applications, including compression artifact reduction \cite{dong2015compression}, image deblurring \cite{sun2015learning} \cite{schuler2016learning}, image deconvolution \cite{xu2014deep}, image denoising \cite{eigen2013restoring}, image inpainting \cite{pathak2016context}, removing dirt/rain noise \cite{jain2009natural}, edge-aware filters \cite{xu2015deep}, image colorization \cite{zhang2016colorful}, and in image segmentation \cite{fang2017automatic}. Recently, CNNs are also used for super-resolution enhancement of images \cite{dong2014learning,kim2016accurate,kim2016deeply,johnson2016perceptual}. Although these single-frame super-resolution methods can be directly applied to light field perspective images to improve their spatial resolution, we expect better performance if the angular information available in the light field data is also exploited. 
  
\begin{figure}
   \includegraphics[width=0.45\textwidth]{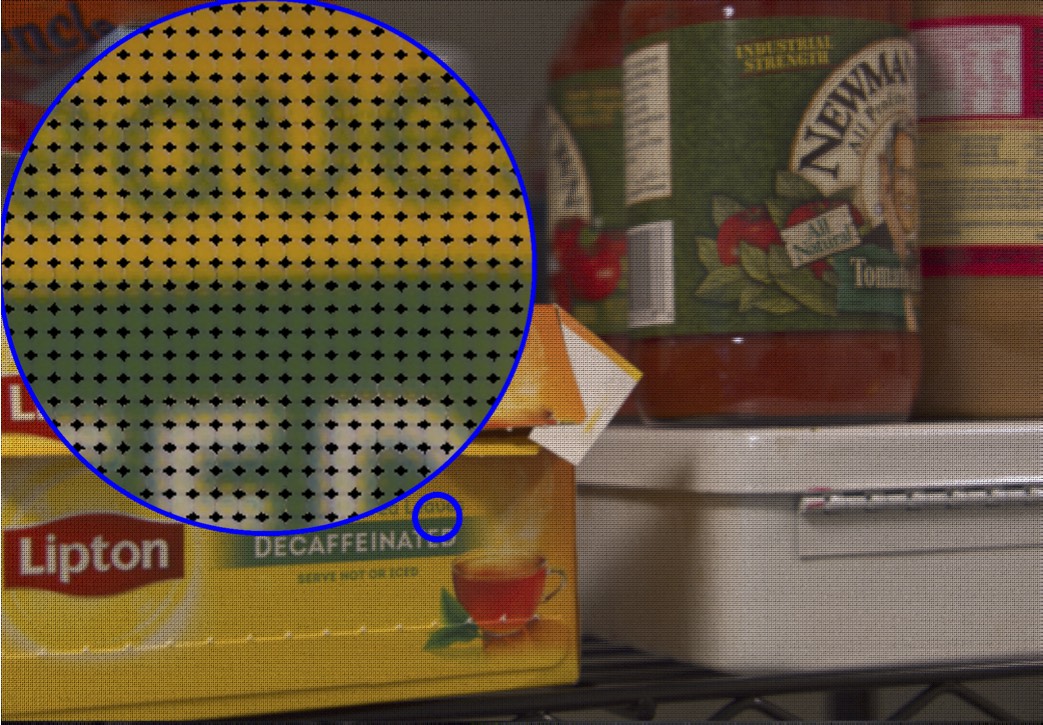}
      \caption{Light field captured by a Lytro Illum camera. A zoomed-in region is overlaid to show the individual lenslet regions.}
      \label{lenslet}
\end{figure}

\begin{figure}
   \includegraphics[width=0.45\textwidth]{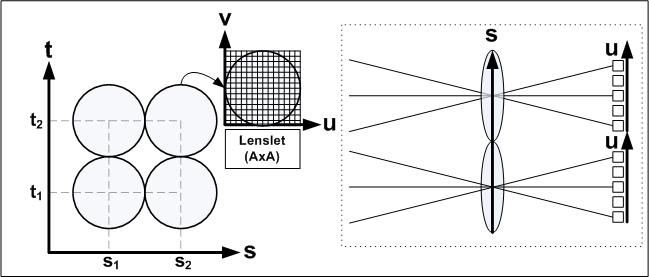}
      \caption{Light field parameterization. Light field can be parameterized by the lenslet positions \textit{(s,t)} and the pixel positions \textit{(u,v)} behind a lenslet.}
      \label{Lensletillustration1}
\end{figure}

\begin{figure}
   \includegraphics[width=0.45\textwidth]{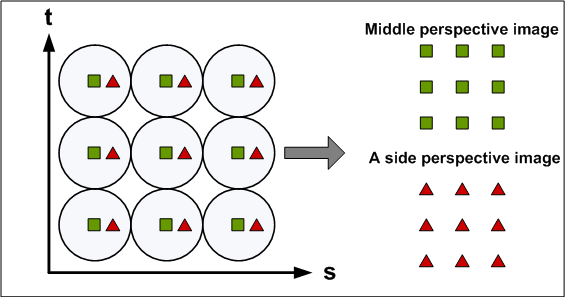}
      \caption{Sub-aperture (perspective) image formation. A perspective image can be constructed by picking specific pixels from the lenslet regions. The size of a perspective image is determined by the number of lenslets.}
      \label{Lensletillustration}
\end{figure}

\section{Light Field Super Resolution using Convolutional Neural Network}

\begin{figure*}
\centering\includegraphics[scale=0.35]{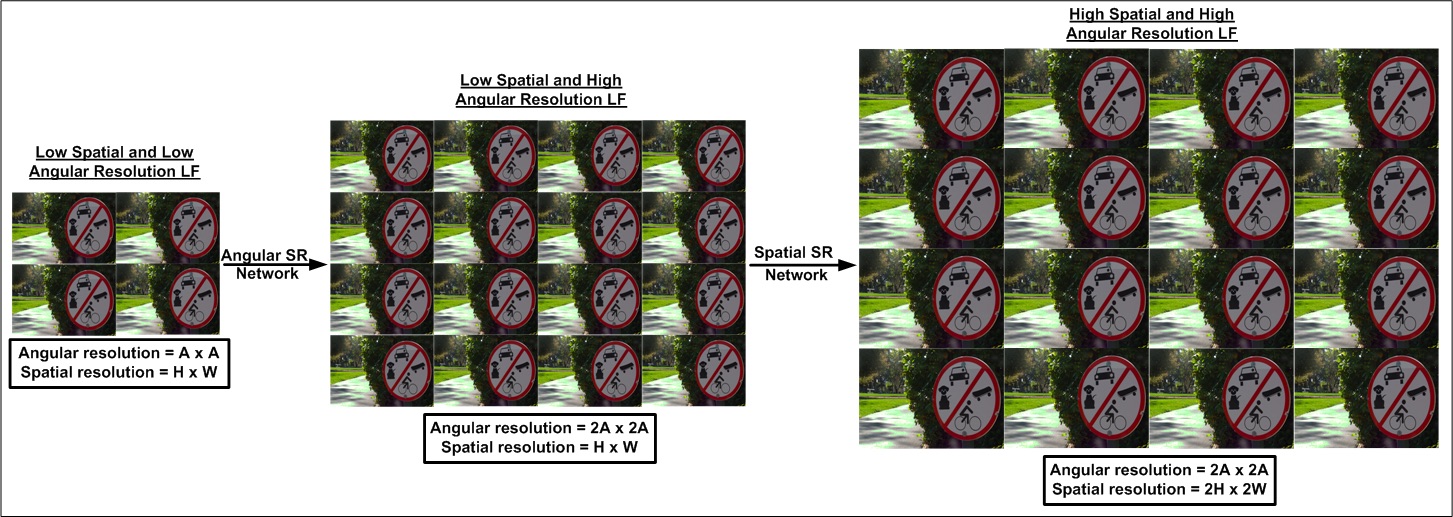}
\caption{An illustration of the proposed LFSR method. First, the angular resolution of the light field (LF) is doubled; second, the spatial resolution is doubled. The networks are applied directly on the raw demosaicked light field, not on the perspective images.}
\label{architecture}
\end{figure*}

In Figure \ref{lenslet}, a light field captured by a micro-lens array based light field camera (Lytro Illum) is shown. When zoomed-in, individual lenslet regions of the MLA can be seen. The pixels behind a lenslet region record directional light intensities received by that lenslet. As illustrated in Figure \ref{Lensletillustration1}, it is possible to represent a light field with four parameters \textit{(s,t,u,v)}, where \textit{(s,t)} indicates the lenslet location, and \textit{(u,v)} indicates the angular position behind the lenslet. A perspective image can be constructed by taking a single pixel value with a specific \textit{(u,v)} index from each lenslet. The process is illustrated in Figure \ref{Lensletillustration}. The spatial resolution of a perspective image is controlled by the size and the number of the lenslets. Given a fixed image sensor size, the spatial resolution can be increased by having smaller size lenslets; given a fixed lenslet size, the spatial resolution can be increased by increasing the number of lenslets, thus, the size of the image sensor. The angular resolution, on the other hand, is defined by the number of pixels behind a lenslet region.     

Our goal is to increase both spatial and angular resolution of a light field capture. We propose a convolutional neural network based learning method, which we call {\it light field super resolution} (LFSR). It consists of two steps. Given a light field where there are $A$x$A$ pixels in each lenslet area and the size of each perspective is $H$x$W$, the first step doubles the angular resolution from $A$x$A$ to $2A$x$2A$ using a convolutional neural network. In the second step, the spatial resolution is doubled from $H$x$W$ to $2H$x$2W$ by estimating new lenslet regions between given lenslet regions. Figure \ref{architecture} gives an illustration of these steps.  

The closest work in the literature to our method is the one presented in \cite{yoon2015learning}, which also uses deep convolutional networks. There is a fundamental difference between our approach and the one in \cite{yoon2015learning}; while our architecture is designed to work on raw light field data, that is, lenslet regions; \cite{yoon2015learning} is designed to work on perspective images. In the experimental results section, we provide both visual and quantitative comparisons with \cite{yoon2015learning}.

\subsection{Angular Super-Resolution (SR) Network}

\begin{figure*}
\centering\includegraphics[scale=0.39]{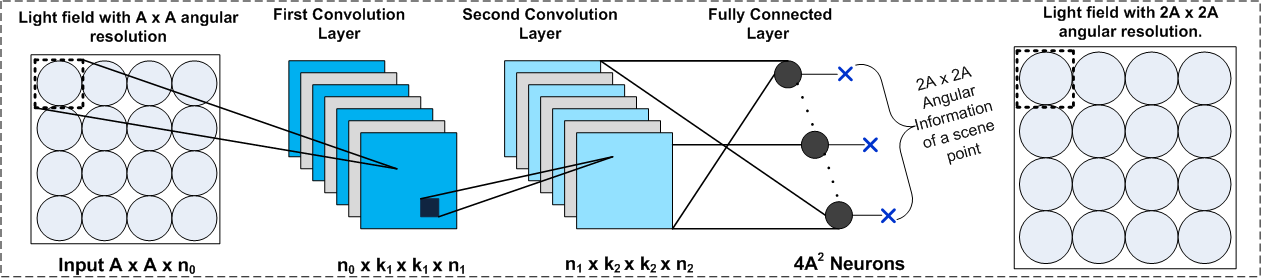}
\caption{Overview of the angular SR network to estimate a higher-angular resolution version of the input light field. A lenslet is drawn as a circle; the $A$x$A$ region behind a lenslet is taken as the input and processed to predict the corresponding $2A$x$2A$ lenslet region. Each convolution layer is followed by a non-linear activation layer of ReLU.}
\label{architectureAR}
\end{figure*}

The proposed angular super-resolution network is shown in Figure \ref{architectureAR}. It is composed of two convolution layers and a fully connected layer. The input to the network is a lenslet region with size $A$x$A$; and the output is a higher resolution lenslet region with size $2A$x$2A$. That is, the angular resolution enhancement is done directly on the raw light field (after demosaicking) as opposed to doing on perspective images. Each lenslet region is interpolated by applying the same network. Once the lenslet regions are interpolated, one can construct the perspective images by rearranging the pixels, as mentioned before. At the end, $2A$x$2A$ perspective images are obtained from $A$x$A$ perspective images.    

The convolution layers in the proposed architecture are based on the intuition that the first layer extracts a high-dimensional feature vector from the lenslet and the second convolution layer maps it onto another high-dimensional vector. After each convolution layer, there is a non-linear activation layer of {\it Rectified Linear Unit} (ReLU). In the end, a fully connected layer aggregates the information of the last convolution layer and predicts a higher-resolution version of the lenslet region. 

The first convolution layer has $n_1$ filters, each with size $n_0$x$k_1$x$k_1$. (In our experiments, we treat each color channel separately, thus $n_0=1$.) The second convolution layer has $n_2$ filters, each with size $n_1$x$k_2$x$k_2$. The final layer is a fully connected layer with $4A^2$ neurons, forming a $2A$x$2A$ lenslet region.

\subsection{Spatial Super-Resolution (SR) Network}

\begin{figure*}
\centering\includegraphics[scale=0.37]{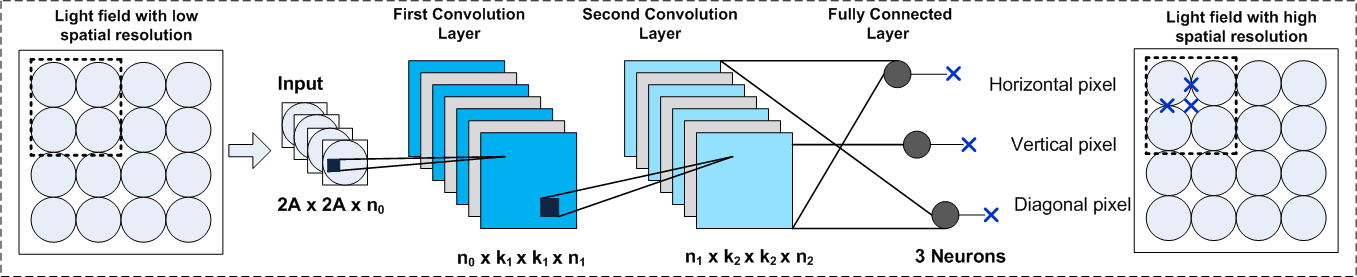}
\caption{Overview of the proposed spatial SR network to estimate a higher-spatial resolution version of the input light field. Four lenslet regions are stacked and given as the input to the network. The network predicts three new pixels to be used in the high-resolution perspective image formation. Each convolution layer is followed by a non-linear activation layer of ReLU.}
\label{architectureSR}
\end{figure*}

\begin{figure}[!htbp]
   \includegraphics[width=0.45\textwidth]{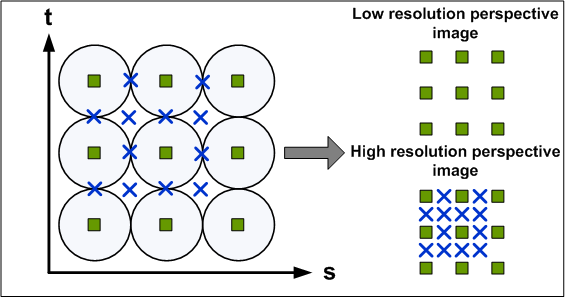}
      \caption{Constructing a high-resolution perspective image. A perspective image can be formed by picking a specific pixel from each lenslet region, and putting all picked pixels together. Using the additional pixels predicted by the spatial SR network, a higher-resolution perspective image is formed.}
      \label{illustration1}
\end{figure}

Figure \ref{architectureSR} gives an illustration of the spatial super-resolution network. Similar to the angular super-resolution network, the architecture has two convolution layers, each followed by a ReLU layer, followed by a fully connected layer. Different from the angular resolution network, four lenslet regions are stacked together as the input to the network. There are three outputs at the end, predicting the horizontal, vertical, and diagonal sub-pixels of a perspective image. To clarify the idea further, Figure \ref{illustration1} illustrates the formation of a high-resolution perspective image. As mentioned earlier, a perspective image of a light field is formed by picking a specific pixel from each lenslet region and putting all picked pixels together according to their respective lenslet positions. Using four lenslet regions, the network predicts three additional pixels in between the pixels picked from the lenslet regions. The predicted pixels, along with the picked pixels, form a higher resolution perspective image.

\subsection{Training the Networks}

We used a dataset that is captured by a Lytro Illum camera \cite{rajlight}. The dataset has more than 200 raw light fields, each with an angular resolution of 14x14 and a spatial resolution of 374x540. In other words, each light field consists of 14x14 perspective images; and each perspective image has a spatial resolution of 374x540 pixels. The raw light field is of size 5236x7560, consisting of 374x540 lenslet regions, where each lenslet region has 14x14 pixels. We used 45 light fields for training and reserved the others for testing. The training data is obtained in two steps. First, we drop every other lenslet region to obtain a low-spatial-resolution (187x270) and high-angular-resolution (14x14) light field. Second, we drop every other pixel in a lenslet region to obtain a low-spatial-resolution (187x270) and low-angular-resolution (7x7) light field. 

\begin{table*}
\caption{Comparison of different spatial and angular resolution enhancement methods.}
\centering
\begin{tabular}{ |p{3cm}|p{1.8cm}|p{1.8cm}|p{1.8cm}|p{1.8cm}|p{1.8cm}|p{1.8cm}|}
 \hline
 \multicolumn{1}{|c}{\multirow{2}{5em}{Methods}}  & \multicolumn{3}{|c}{PSNR (dB)}  & \multicolumn{3}{|c|}{SSIM}\\ 
\cline{2-7}
        &Min&Avg&Max&Min&Avg&Max \\
 \hline
  Bicubic resizing (\textit{imresize}) & 24.2029 & 27.6671 & 34.6330 & 0.7869 & 0.8744 & 0.9457  \\ 
   \hline
 LFCNN \cite{yoon2015learning} & 25.5963 & 28.9661 & 34.8231 & 0.7838 & 0.8904 & 0.9407 \\
 \hline
 Bicubic interpolation & 27.2620 & 30.6245 & 37.1640 & 0.5780 & 0.9256 & 0.9659  \\
 \hline
Proposed (LFSR) & \redbf{29.7515} & \redbf{33.4273} & \redbf{39.5655} & \redbf{0.9360} & \redbf{0.9559} & \redbf{0.9823}\\
 \hline 
 \end{tabular}
\label{table:psnr}
\end{table*}

The angular SR network, as shown in Figure \ref{architectureAR}, has low-spatial-resolution and low-angular-resolution light field as its input, and low-spatial-resolution and high-angular-resolution light field as its output. Each lenslet region is treated separately by the network, increasing the size from 7x7 to 14x14. The first convolution layer consists of 64 filters, each with size 1x3x3. It is followed by a ReLU layer. The second convolution layer consists of 32 filters of size 64x1x1, followed by a ReLU layer. Finally, there is a fully connected layer with 196 neurons to produce a 14x14 lenslet region.  

The spatial SR network, as shown in Figure \ref{architectureSR}, has low-spatial-resolution and high-angular-resolution light field as its input, and high-spatial-resolution and high-angular-resolution light field as its output. Four lenslet regions are stacked to form a 14x14x4 input. The first convolution layer consists of 64 filters, each with size 4x3x3. The second convolution layer consists of 32 filters of size 64x1x1. Each convolution layer is followed by a ReLU layer. Finally, there is a fully connected layer with three neurons to produce the horizontal, vertical and diagonal pixels. This network generates one high-spatial resolution perspective. For each perspective, the network is trained separately.  

We implement and train our model using the Caffe package \cite{jia2014caffe}. For the weight initialization of both networks, we used the initialization technique given in \cite{glorot2010understanding}, with mean value set to zero and standard deviation set to $10^{-3}$, to prevent vanishment or over-amplification of weights. The learning rates for the three layers of the networks are $10^{-3}$, $10^{-3}$, and $10^{-5}$, respectively. Mean squared error is used as the loss function, which is minimized using the stochastic gradient descent method with standard backpropagation \cite{lecun1998gradient}. For each network, the input size is about 13 million; and the number of iterations is about $10^8$.

\section{Experiments}

We evaluated our LFSR method on 25 test light fields which we reserved from the Lytro Illum camera dataset \cite{rajlight} and on the HCI dataset \cite{wanner2013datasets}. For spatial and angular resolution enhancement, we compared our method against the LFCNN \cite{yoon2015learning} method and bicubic interpolation. There are several methods in the literature that synthesize new viewpoints from a light field data; thus, we compared the angular SR network of our method with two such view synthesis methods, namely, Kalantari \textit{et al.} \cite{LearningViewSynthesis} and Wanner and Goldluecke \cite{wanner2014variational}. Finally, there are single-frame spatial resolution enhancement methods; we chose the latest state-of-the-art method, called DRRN \cite{Tai-DRRN-2017}, and included it in our comparisons. 

In addition to spatial and angular resolution enhancement, we investigated depth estimation performance, and compared the depth maps generated by low-resolution light fields and the resolution-enhanced light fields. In the end, we investigated the effect of the network parameters, including the filter size and the number of layers, on the performance of the proposed spatial SR network.

\subsection{Spatial and Angular Resolution Enhancement}

The test images are downsampled from 14x14 perspective images, each with size 374x540 pixels, to 7x7 perspective images with size 187x270 pixels by dropping every other lenslet region and every pixel in each lenslet region. The trained networks are applied to these low-spatial and low-angular resolution images to bring them back to the original spatial and angular resolutions. The networks are applied on each color channel separately. Since the original perspective images available, we can quantitatively calculate the performance by comparing the estimated and the original images.  In Table \ref{table:psnr}, we provide peak-signal-to-noise ratio (PSNR) and structural similarity index (SSIM) \cite{wang2004image} results of our method, in addition to the results of the LFCNN \cite{yoon2015learning} method and bicubic interpolation. Here, we should make two notes about the LFCNN method. First, we took the learned parameters provided in the original paper and fine tuned them with our dataset as described in \cite{yoon2015learning}. This revision improves the performance of the LFCNN method for our dataset. Second, the LFCNN method is designed to split a low-resolution image pixel into four sub-pixels to produce a high-resolution image; therefore, we included the results of bicubic resizing (\textit{imresize} function in MATLAB) to evaluate the quantitative performance of the LFCNN method. In Table \ref{table:psnr}, we see that the LFCNN method produces about 1.3 dB better than the bicubic resizing. The proposed method produces the best results in terms of PSNR and SSIM. 

\begin{figure*}
\centering
\captionsetup{justification=centering,margin=0cm}
  {\includegraphics[width=.5\textwidth]{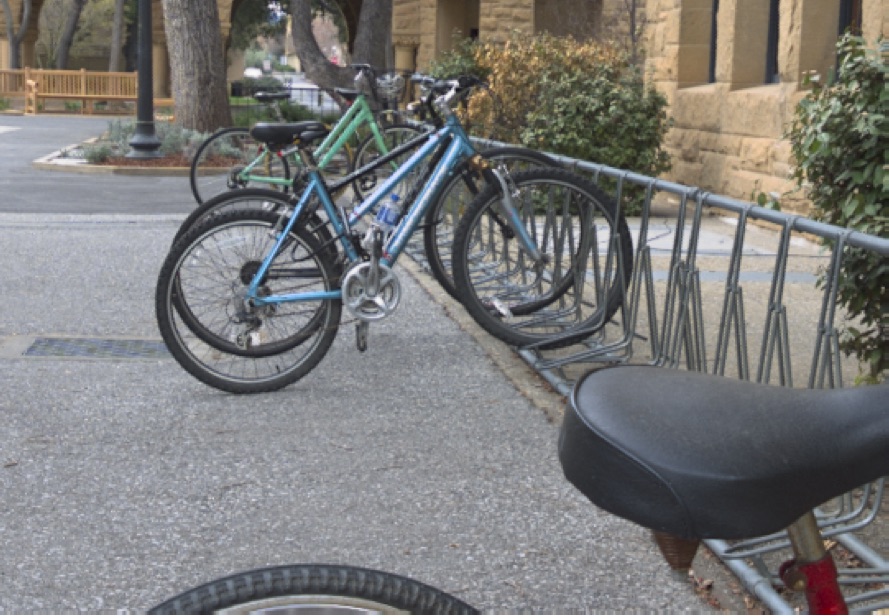}}\vspace{2mm}
 
  {\includegraphics[trim= 75 125 250 100, clip=true, width=0.16\textwidth]{GT9}}\hfill
  {\includegraphics[trim= 75 125 250 100, clip=true, width=0.16\textwidth]{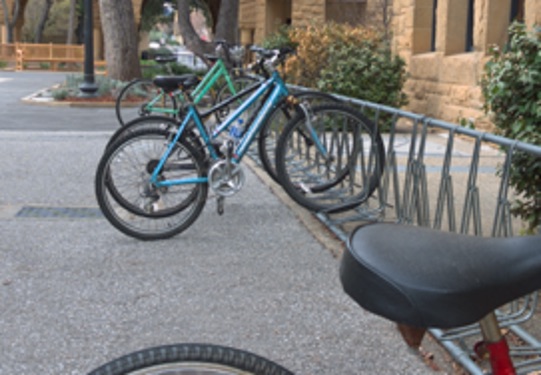}}\hfill
  {\includegraphics[trim= 75 125 250 100, clip=true, width=0.16\textwidth]{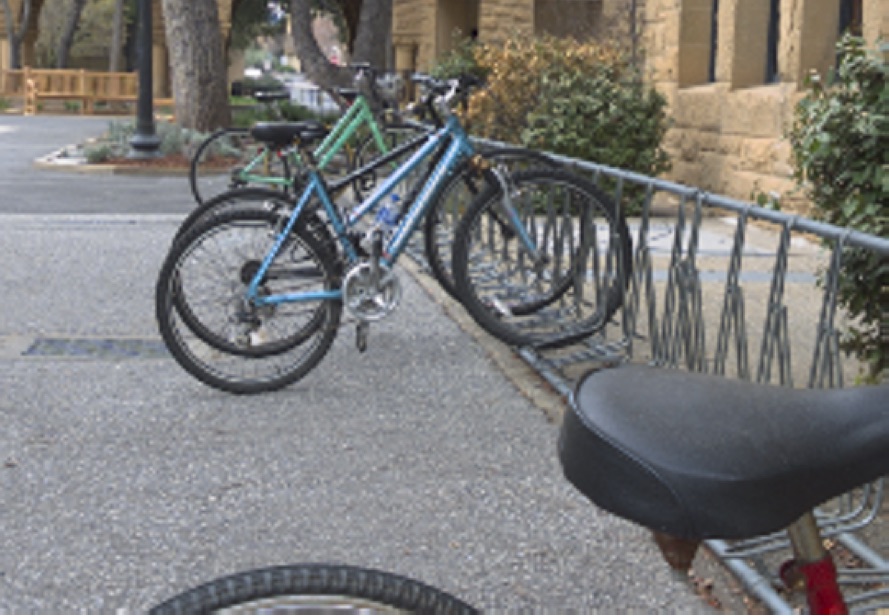}}\hfill
  {\includegraphics[trim= 75 125 250 100, clip=true, width=0.16\textwidth]{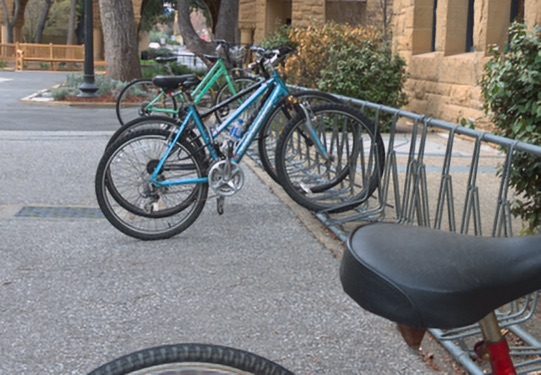}}\hfill
  {\includegraphics[trim= 95 155 312 126, clip=true, width=0.16\textwidth]{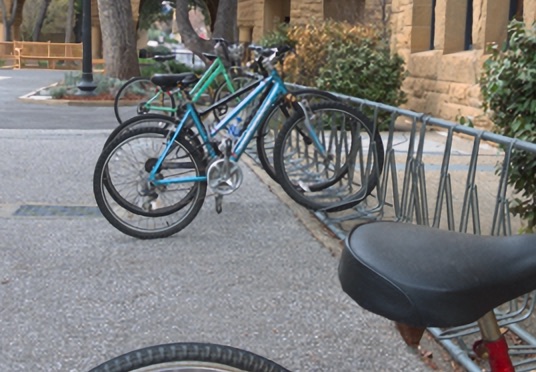}}\hfill
  {\includegraphics[trim= 75 125 250 100, clip=true, width=0.16\textwidth]{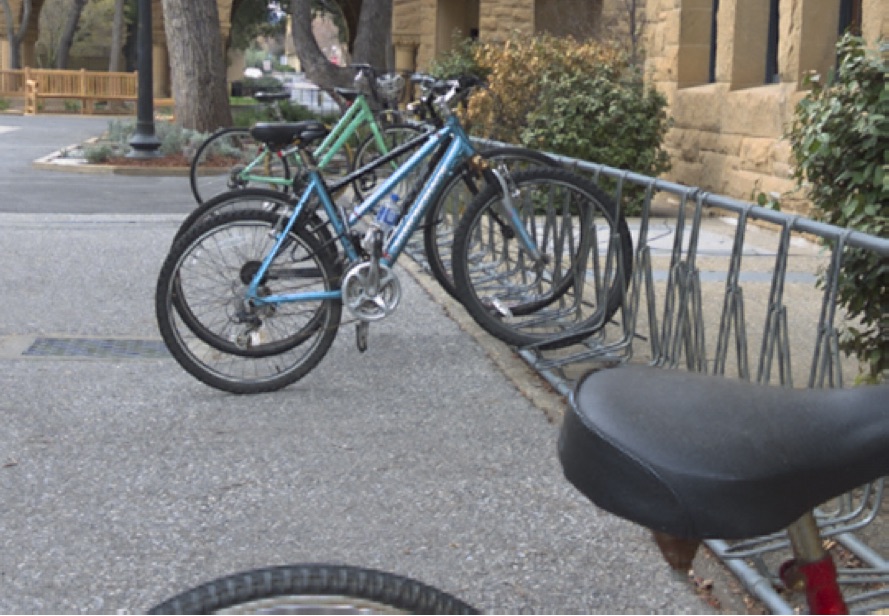}}\hfill
  \subfloat[Ground truth\label{fig:test11}]
  {\includegraphics[trim= 250 123 0 50, clip=true, width=0.16\textwidth]{GT9}}\hfill
  \subfloat[Bicubic resizing (\textit{imresize}) / 25.34 dB\label{fig:test12}]
  {\includegraphics[trim= 250 123 0 50, clip=true, width=0.16\textwidth]{Bicubic_new9}}\hfill
  \subfloat[Bicubic interp. / 27.7 dB\label{fig:test13}]
  {\includegraphics[trim= 250 123 0 50, clip=true, width=0.16\textwidth]{B9}}\hfill
    \subfloat[LFCNN \cite{yoon2015learning} / 25.76 dB\label{fig:test14}]
  {\includegraphics[trim= 250 123 0 50, clip=true, width=0.16\textwidth]{LFCNN9}}\hfill
    \subfloat[DRRN \cite{Tai-DRRN-2017} / 31.63 dB\label{fig:testDRRN1}]
  {\includegraphics[trim= 320 152 0 69, clip=true, width=0.16\textwidth]{2_x2}}\hfill
  \subfloat[LFSR / 32.25 dB\label{fig:test15}]
  {\includegraphics[trim= 250 123 0 50, clip=true, width=0.16\textwidth]{P9}}\hfill
\caption{Visual comparison of different methods.}
\label{fig:test1}
\end{figure*}

\begin{figure*}
\centering
\captionsetup{justification=centering,margin=0cm}
  {\includegraphics[width=.5\textwidth]{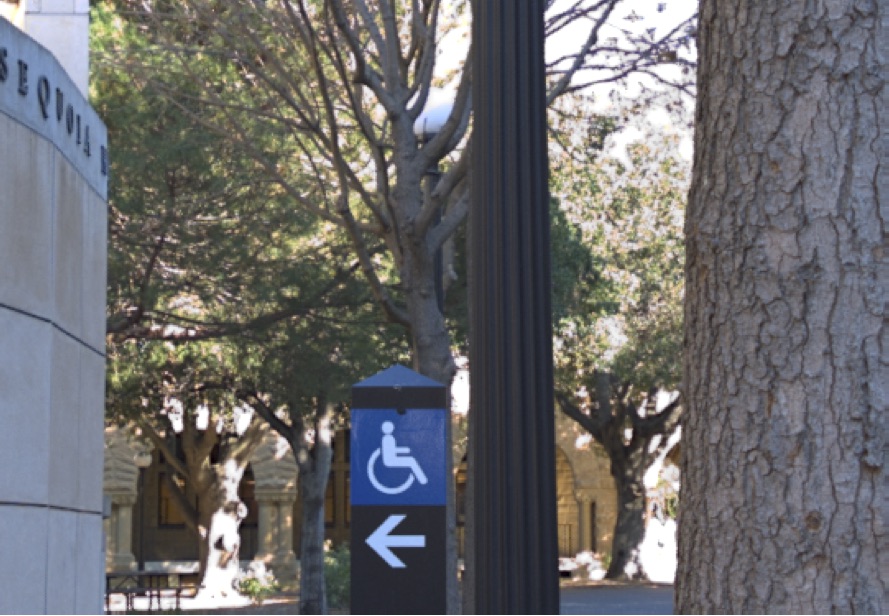}}\vspace{2mm}

  {\includegraphics[trim= 170 55 220 200, clip=true, width=0.16\textwidth]{GT11}}\hfill
  {\includegraphics[trim= 170 56 220 200, clip=true, width=0.16\textwidth]{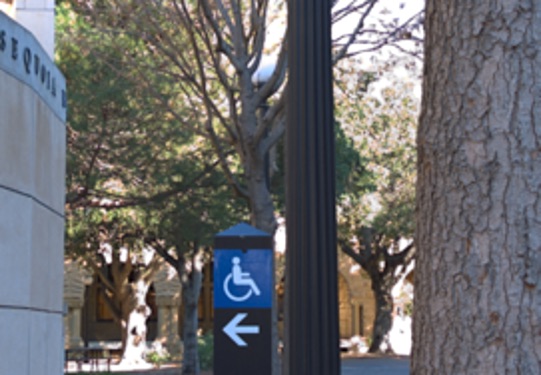}}\hfill
  {\includegraphics[trim= 170 55 220 200, clip=true, width=0.16\textwidth]{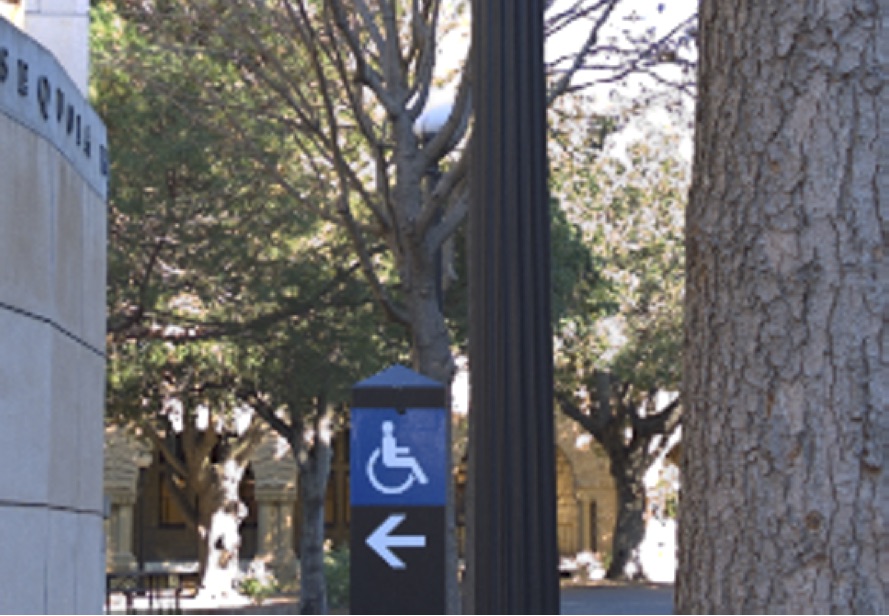}}\hfill
  {\includegraphics[trim= 170 56 220 200, clip=true, width=0.16\textwidth]{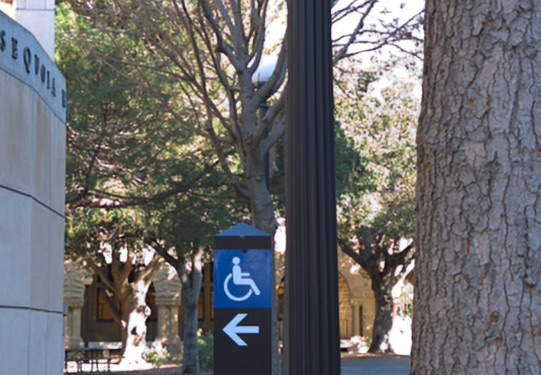}}\hfill
  {\includegraphics[trim= 215 68 274 252, clip=true, width=0.16\textwidth]{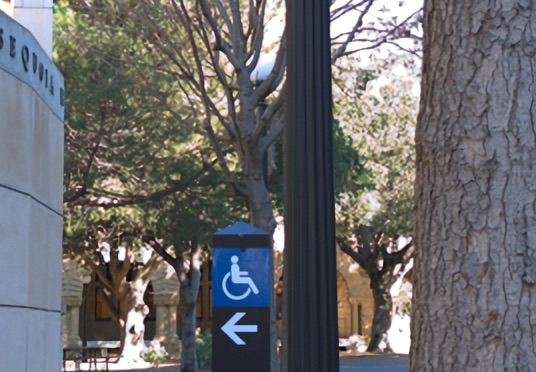}}\hfill
  {\includegraphics[trim= 170 55 220 200, clip=true, width=0.16\textwidth]{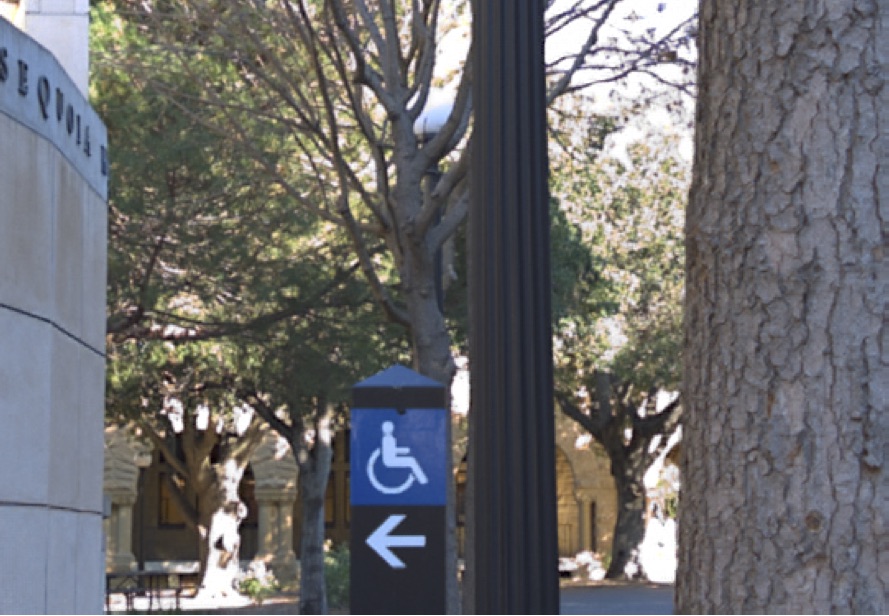}}\hfill
  
\subfloat[Ground truth\label{fig:test31}]
 {\includegraphics[trim= 0 220 370 20, clip=true, width=0.16\textwidth]{GT11}}\hfill
\subfloat[Bicubic resizing (\textit{imresize}) / 25 dB\label{fig:test32}]
  {\includegraphics[trim= 0 221 370 20, clip=true, width=0.16\textwidth]{Bicubic_new11}}\hfill
\subfloat[Bicubic interp. / 28.11 dB\label{fig:test33}]
  {\includegraphics[trim= 0 220 370 20, clip=true, width=0.16\textwidth]{B11}}\hfill
  \subfloat[LFCNN \cite{yoon2015learning} / 25.12 dB\label{fig:test34}]
  {\includegraphics[trim= 0 221 370 20, clip=true, width=0.16\textwidth]{LFCNN11}}\hfill
  \subfloat[DRRN \cite{Tai-DRRN-2017} / 32.20 dB\label{fig:test35}]
  {\includegraphics[trim= 0 277 465 25, clip=true, width=0.16\textwidth]{24_x2}}\hfill
\subfloat[LFSR / 32.35 dB\label{fig:test36}]
  {\includegraphics[trim= 0 220 370 20, clip=true, width=0.16\textwidth]{P11}}\hfill
  
\caption{Visual comparison of different methods.}
\label{fig:test2}
\end{figure*}

\begin{figure*}
\centering
\captionsetup{justification=centering,margin=0cm}
  {\includegraphics[width=.5\textwidth]{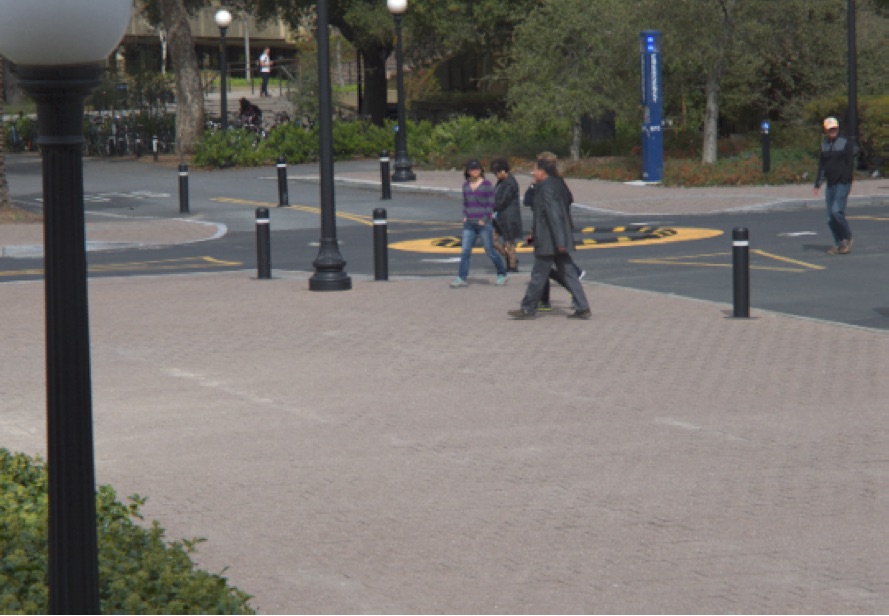}}\vspace{2mm}

  {\includegraphics[trim= 110 163 280 98, clip=true, width=0.16\textwidth]{GT21}}\hfill
  {\includegraphics[trim= 110 164 280 98, clip=true, width=0.16\textwidth]{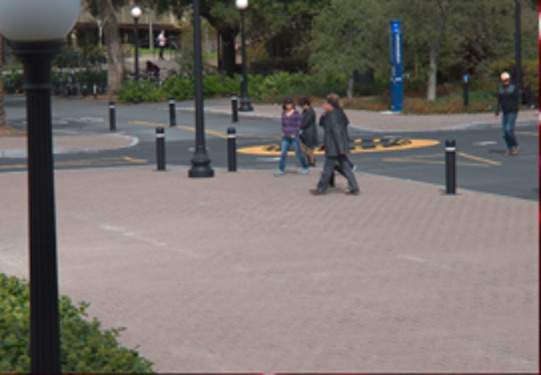}}\hfill
  {\includegraphics[trim= 110 163 280 98, clip=true, width=0.16\textwidth]{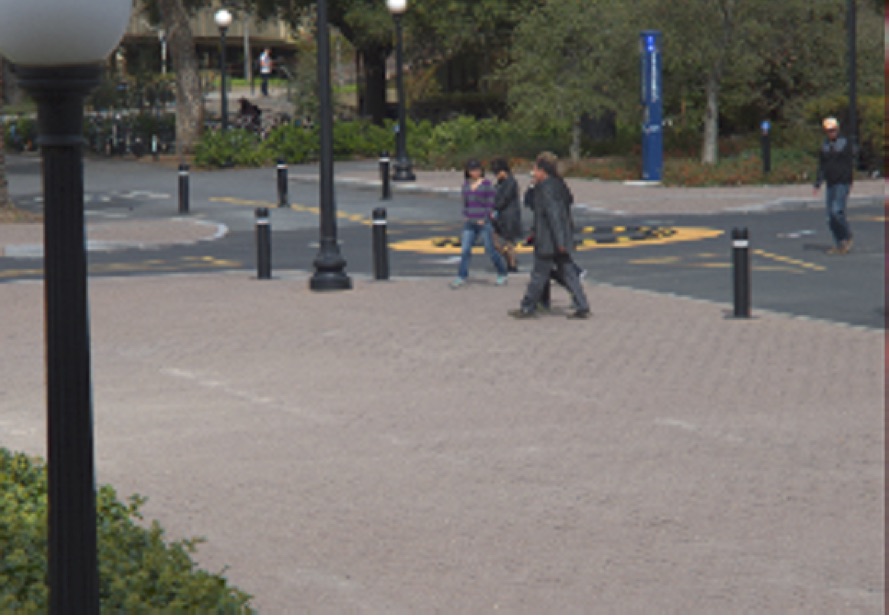}}\hfill
  {\includegraphics[trim= 110 164 280 98, clip=true, width=0.16\textwidth]{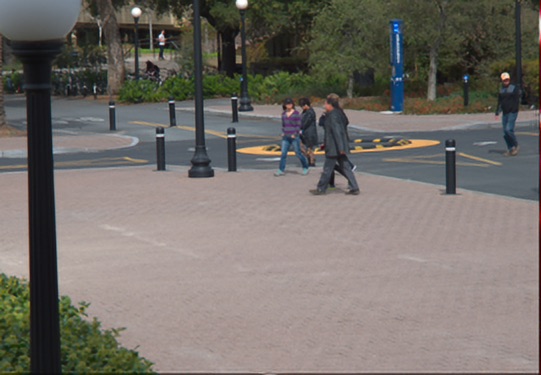}}\hfill
  {\includegraphics[trim= 140 206 351 123, clip=true, width=0.16\textwidth]{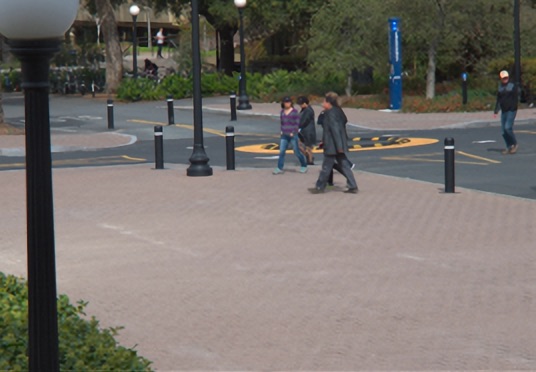}}\hfill
  {\includegraphics[trim= 110 163 280 98, clip=true, width=0.16\textwidth]{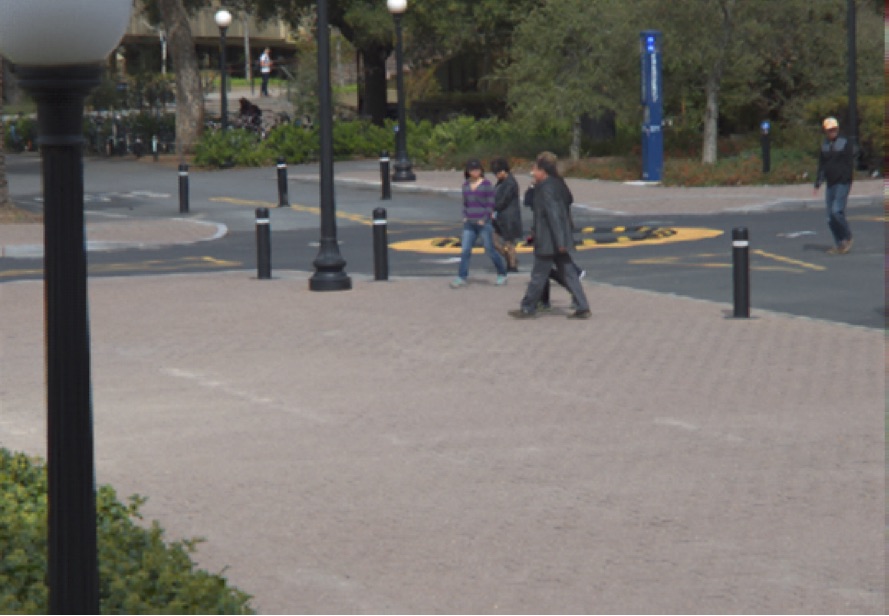}}\hfill
\subfloat[Ground truth\label{fig:test51}]
  {\includegraphics[trim= 275 170 125 100, clip=true, width=0.16\textwidth]{GT21}}\hfill
\subfloat[Bicubic resizing (\textit{imresize}) / 28.92 dB\label{fig:test52}]
  {\includegraphics[trim= 275 171 125 100, clip=true, width=0.16\textwidth]{Bicubic_new21}}\hfill
\subfloat[Bicubic interp. / 28.11 dB\label{fig:test53}]  
  {\includegraphics[trim= 275 170 125 100, clip=true, width=0.16\textwidth]{B21}}\hfill
  \subfloat[LFCNN \cite{yoon2015learning} / 29.06 dB\label{fig:test54}]
  {\includegraphics[trim= 275 171 125 100, clip=true, width=0.16\textwidth]{LFCNN21}}\hfill
  \subfloat[DRRN \cite{Tai-DRRN-2017} / 36.78 dB\label{fig:test55}]
  {\includegraphics[trim= 348 214 154 125, clip=true, width=0.16\textwidth]{39_x2}}\hfill
\subfloat[LFSR / 34.21 dB\label{fig:test56}]
  {\includegraphics[trim= 275 170 125 100, clip=true, width=0.16\textwidth]{P21}}\hfill
\caption{Visual comparison of different methods.}
\label{fig:test3}
\end{figure*}

\begin{figure*}
\centering
\captionsetup{justification=centering,margin=0cm}
  {\includegraphics[width=.5\textwidth]{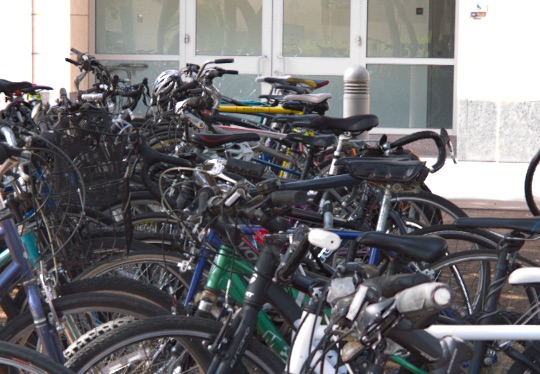}}\vspace{2mm}

  {\includegraphics[trim= 330 240 150 55, clip=true, width=0.16\textwidth]{GT3}}\hfill
  {\includegraphics[trim= 330 241 150 55, clip=true, width=0.16\textwidth]{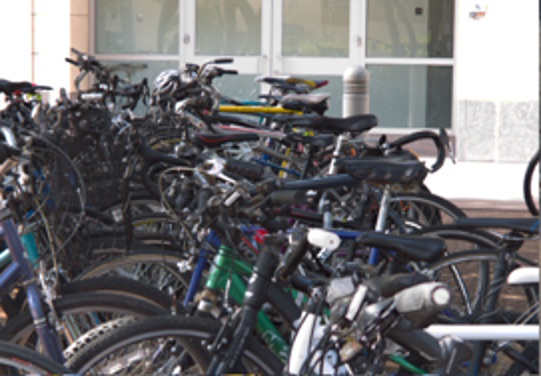}}\hfill
  {\includegraphics[trim= 330 240 150 55, clip=true, width=0.16\textwidth]{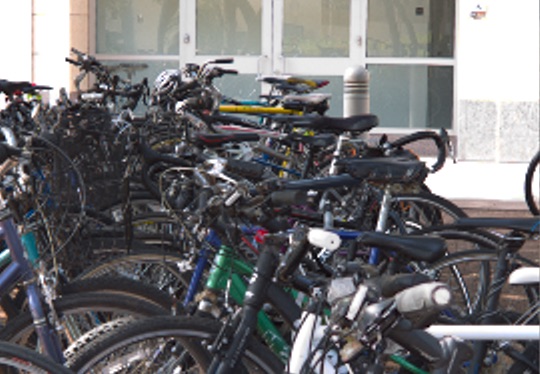}}\hfill
  {\includegraphics[trim= 330 240 150 55, clip=true, width=0.16\textwidth]{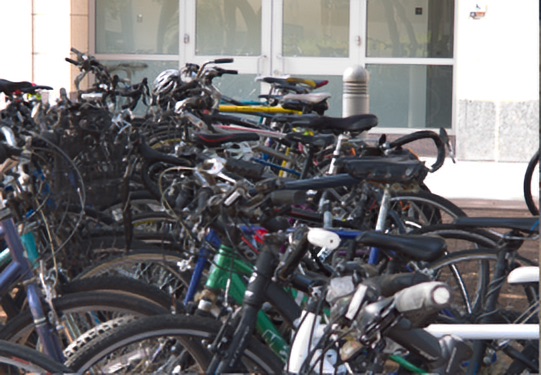}}\hfill
  {\includegraphics[trim= 328 240 148 53, clip=true, width=0.16\textwidth]{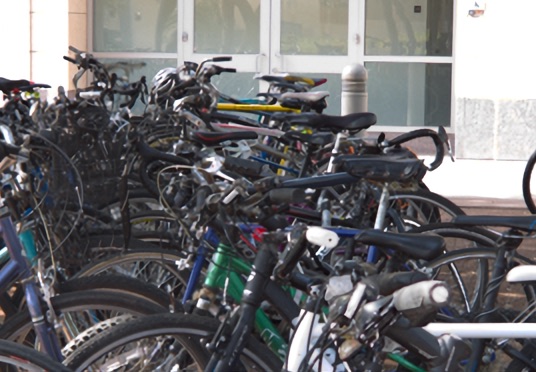}}\hfill
  {\includegraphics[trim= 330 240 150 55, clip=true, width=0.16\textwidth]{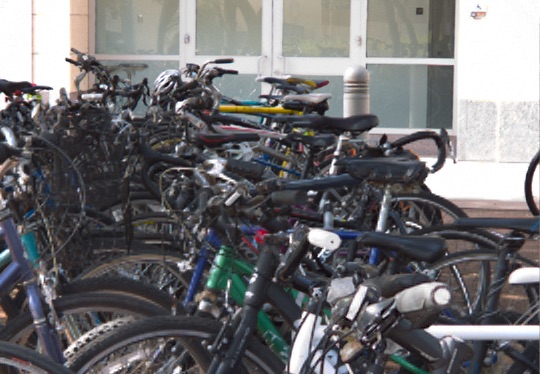}}\hfill
  
\subfloat[Ground truth\label{fig:test91}]
 {\includegraphics[trim= 430 35 50 260, clip=true, width=0.16\textwidth]{GT3}}\hfill
\subfloat[Bicubic resizing (\textit{imresize}) / 25.99 dB\label{fig:test92}]
  {\includegraphics[trim= 430 35 50 260, clip=true, width=0.16\textwidth]{B3_new}}\hfill
\subfloat[Bicubic interp. / 28.15 dB\label{fig:test93}]
  {\includegraphics[trim= 430 35 50 260, clip=true, width=0.16\textwidth]{B3}}\hfill
  \subfloat[LFCNN \cite{yoon2015learning} / 24.09 dB\label{fig:test94}]
  {\includegraphics[trim= 430 35 50 260, clip=true, width=0.16\textwidth]{worst_LFC}}\hfill
  \subfloat[DRRN \cite{Tai-DRRN-2017} / 33.65 dB\label{fig:test95}]
  {\includegraphics[trim= 428 35 50 261, clip=true, width=0.16\textwidth]{worse_image_x2}}\hfill
\subfloat[LFSR / 29.75 dB\label{fig:test96}]
  {\includegraphics[trim= 430 35 50 260, clip=true, width=0.16\textwidth]{P3}}\hfill
  
\caption{Visual comparison of different methods. (The worst result image from the dataset is shown here.)}
\label{fig:worst}
\end{figure*}

Visual comparison is critical when evaluating spatial resolution enhancement. Figures \ref{fig:test1}, \ref{fig:test2}, and \ref{fig:test3} are typical results from the test dataset. Figure \ref{fig:worst} is our worst result among all test images. In these figures, we also include the results of the single-image spatial resolution method, called DRRN \cite{Tai-DRRN-2017}. This method is based on deep recursive residual network technique, and produces state-of-the-art results in spatial resolution enhancement. Examining the results visually, we conclude that our method performs better the LFCNN method and bicubic interpolation, and produces comparable results with the DRNN method. We notice that the LFCNN method produces sharper results compared to bicubic interpolation despite having lower PSNR values. In our worst result, given in Figure \ref{fig:worst}, the DRNN method outperforms all methods. This particular image has highly complex texture, which seems to be not modelled well with the proposed architecture. Training with similar images or using more complex architecture may improve the performance. When comparing deep networks, we should consider the computational cost as well. The computation time for one image with the DRRN method is about 859 seconds, whereas, the proposed SR network takes about 53 seconds, noting that both are implemented in MATLAB on the same machine.

\begin{figure*}
\centering
\captionsetup{justification=centering,margin=0cm}
  {\includegraphics[width=.24\textwidth]{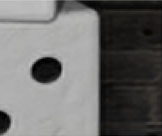}}\hfill
  {\includegraphics[width=.24\textwidth]{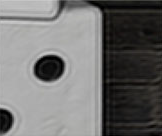}}\hfill
  {\includegraphics[trim= 475 120 200 565, clip=true, width=0.23\textwidth]{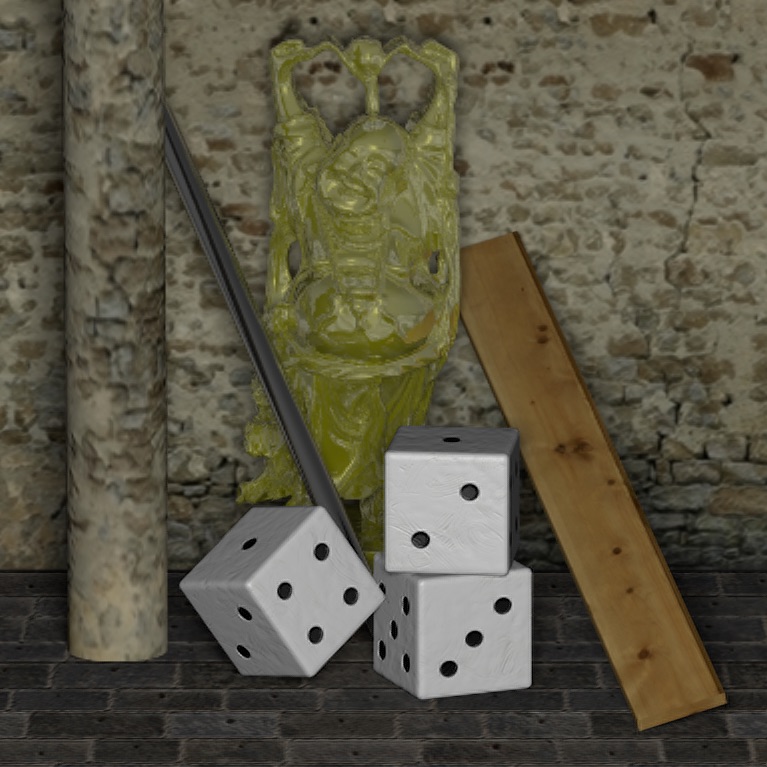}}\hfill
  {\includegraphics[trim= 476 121 200 565, clip=true, width=0.23\textwidth]{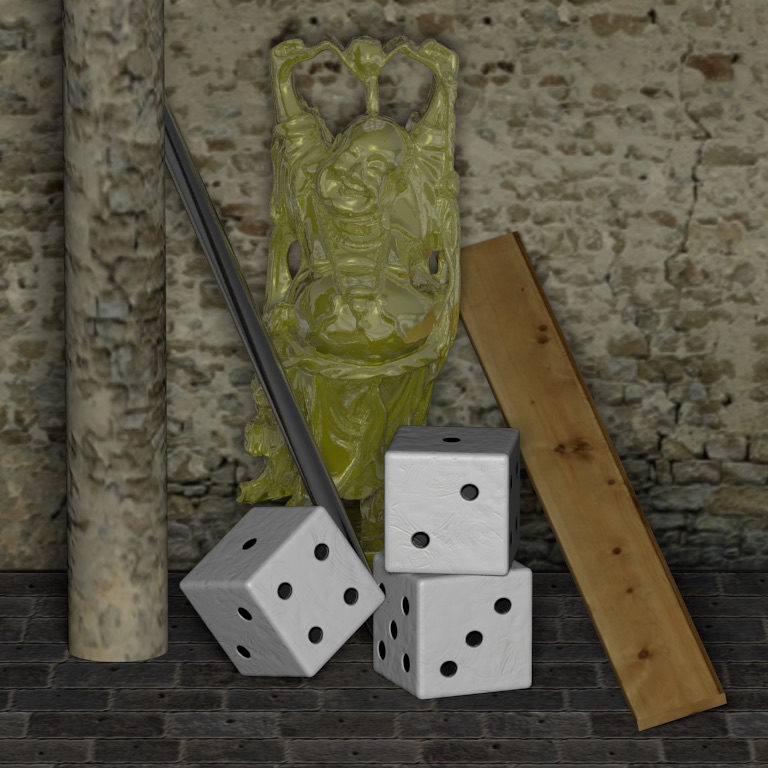}}\hfill
  \subfloat[]
  {\includegraphics[width=.24\textwidth]{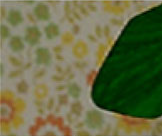}}\hfill
  \subfloat[]
  {\includegraphics[width=.24\textwidth]{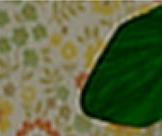}}\hfill
  \subfloat[]
  {\includegraphics[trim= 50 185 628 504, clip=true, width=0.23\textwidth]{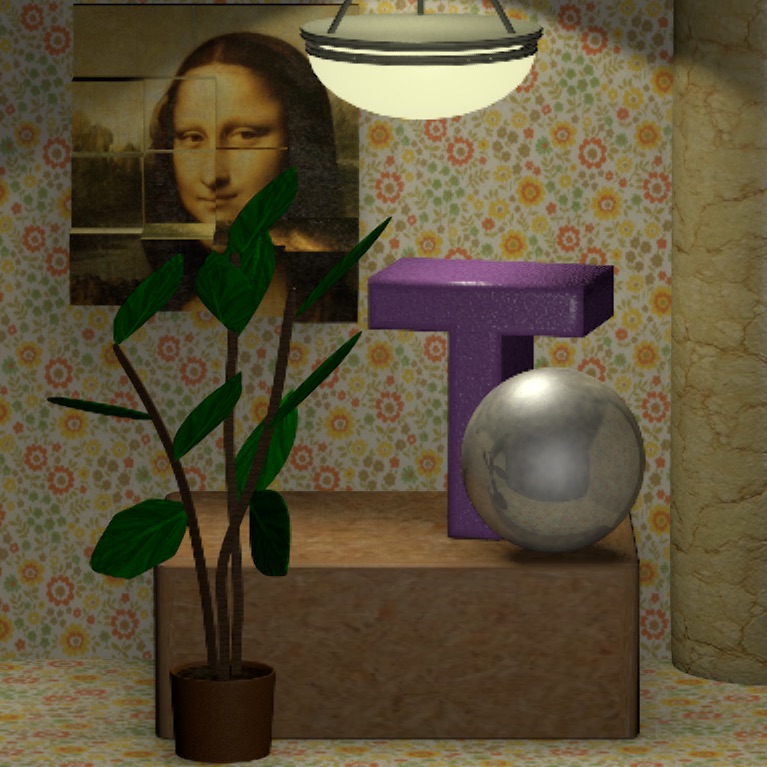}}\hfill
  \subfloat[]
  {\includegraphics[trim= 50 185 628 504, clip=true, width=0.23\textwidth]{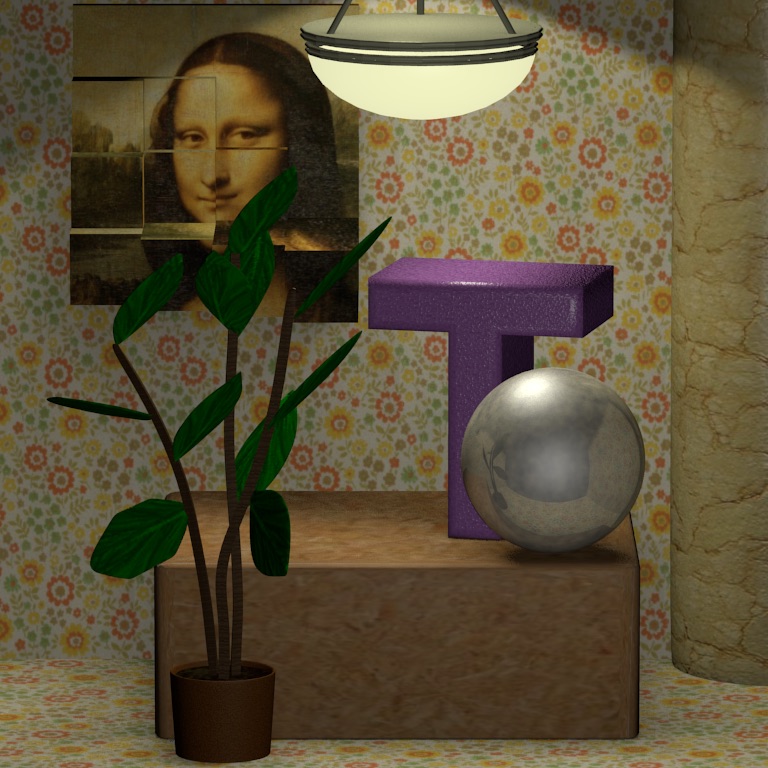}}\hfill
\caption{Visual comparison of different methods for generating novel views from the HCI dataset. (a) Yoon et al. \cite{yoon2017light}. (b) Yoon et al. \cite{yoon2015learning}. (c) Proposed LFSR. (d) Ground truth.}
\label{fig:testXX}
\end{figure*}

In Figure \ref{fig:testXX}, we test our method on the HCI dataset \cite{wanner2013datasets}. We compare against the networks in \cite{yoon2015learning} and \cite{yoon2017light}. The method in \cite{yoon2017light} produces less ringing artifacts compared to the LFCNN network  \cite{yoon2015learning}. The proposed method again produces the best visual results. 

Although we have showed results for resolution enhancement of the middle perspective image so far, the proposed spatial SR network can be used for any perspective image as well. In Table \ref{table:psnr1}, average PSNR and SSIM on test images for different perspective images (among the 14x14 set) are presented. It is seen that similar results are obtained on all perspective images, as expected.

\begin{table}
\caption{Evaluation of the proposed method for different perspective images.}\label{table:psnr1}
\centering
\begin{tabular}{ |p{2.1cm}|p{1.8cm}|p{1.4cm}|p{1.4cm}|}
 \hline
\multirow{2}{8em}{Perspective image (Row \#, Column \#)}  & \multirow{2}{5em}{ Method} \ & \multirow{2}{5em}{PSNR (dB)}  & \multirow{2}{5em}{SSIM} \\ 
& & & \vspace{1mm}\\ 
\hline 
\multirow{2}{8em}{(7,1)}& Bicubic interp. & 28.21 & 0.8631\\ 
\cline{2-4}
& Proposed & \redbf{28.74} & \redbf{0.8789} \\

\hline
\multirow{2}{8em}{(5,5)}& Bicubic interp. & 31.12 & 0.9310\\
\cline{2-4}
& Proposed & \redbf{32.95} & \redbf{0.9496} \\

\hline
\multirow{2}{8em}{(6,6)}& Bicubic interp. & 30.74 & 0.9272 \\
\cline{2-4}
& Proposed & \redbf{32.71} & \redbf{0.9485} \\

\hline
\multirow{2}{8em}{(6,8)}& Bicubic interp. & 30.73 & 0.9267 \\
\cline{2-4}
& Proposed & \redbf{32.94} & \redbf{0.9504} \\

\hline
\multirow{2}{8em}{(8,6)}& Bicubic interp. & 30.69 & 0.9270 \\
\cline{2-4}
& Proposed & \redbf{32.69} & \redbf{0.9492} \\

\hline
\multirow{2}{8em}{(8,8)}& Bicubic interp. & 27.71 & 0.8793 \\
\cline{2-4}
& Proposed & \redbf{28.16} & \redbf{0.8917} \\

 \hline 
 \end{tabular}
\end{table}

\subsection{Angular Resolution Enhancement}
\begin{figure*}
\centering
\captionsetup{justification=centering,margin=0cm}
  {\includegraphics[width=.5\textwidth]{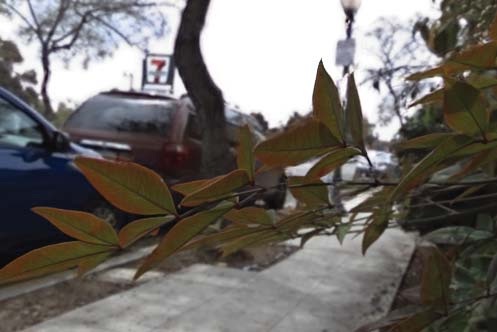}}\vspace{2mm}
 
  {\includegraphics[width=0.12\textwidth]{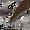}}\hfill
  {\includegraphics[width=0.12\textwidth]{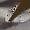}}\hfill
  {\includegraphics[width=0.12\textwidth]{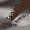}}\hfill
  {\includegraphics[width=0.12\textwidth]{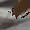}}\hfill
  {\includegraphics[width=0.12\textwidth]{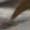}}\hfill
  {\includegraphics[trim= 143 63 367 282, clip=true,width=0.12\textwidth]{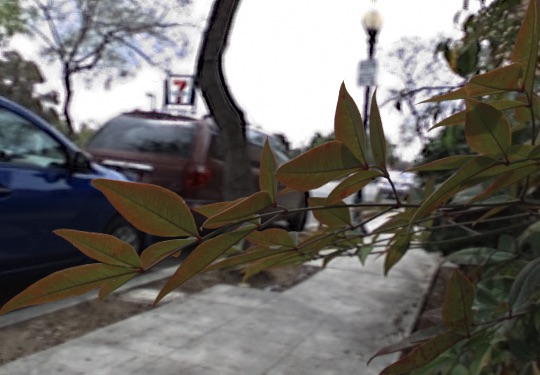}}\hfill
  {\includegraphics[width=0.12\textwidth]{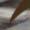}}\vspace{2mm}
  
  {\includegraphics[width=0.12\textwidth]{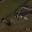}}\hfill
  {\includegraphics[width=0.12\textwidth]{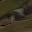}}\hfill
  {\includegraphics[width=0.12\textwidth]{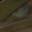}}\hfill
  {\includegraphics[width=0.12\textwidth]{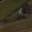}}\hfill
  {\includegraphics[width=0.12\textwidth]{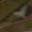}}\hfill
  {\includegraphics[trim= 218 109 292 236, clip=true,width=0.12\textwidth]{Ravi_AR1_4}}\hfill
  {\includegraphics[width=0.12\textwidth]{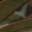}}
 
  \subfloat[15.03 dB\label{fig:ravi1}]
  {\includegraphics[width=0.12\textwidth]{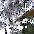}}\hfill
  \subfloat[23.54 dB\label{fig:ravi2}]
  {\includegraphics[width=0.12\textwidth]{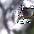}}\hfill
  \subfloat[23.88 dB\label{fig:ravi3}]
  {\includegraphics[width=0.12\textwidth]{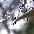}}\hfill
  \subfloat[20.08 dB\label{fig:ravi4}]
  {\includegraphics[width=0.12\textwidth]{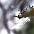}}\hfill
  \subfloat[27.80 dB\label{fig:ravi5}]
  {\includegraphics[width=0.12\textwidth]{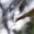}}\hfill
  \subfloat[33.08 dB\label{fig:ravi6}]
  {\includegraphics[trim= 413 234 95 109, clip=true,width=0.12\textwidth]{Ravi_AR1_4}}\hfill
 \subfloat[Ground truth\label{fig:ravi8}]
  {\includegraphics[width=0.12\textwidth]{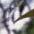}}\hfill
  
\caption{Visual comparison of different methods for novel view synthesis. The picture (\textit{"Leaves"}) is taken from Kalantari \textit{et al.} \cite{LearningViewSynthesis}.  (a) \cite{wanner2014variational} with disparity \cite{wanner2012globally}. (b) \cite{wanner2014variational} with disparity \cite{Tao_2013_ICCV}. (c) \cite{wanner2014variational} with disparity \cite{wang2015occlusion}. (d) \cite{wanner2014variational} with disparity \cite{jeon2015accurate}. (e) Kalantari \textit{et al.} \cite{LearningViewSynthesis}. (f) Proposed angular SR network. (g) Ground truth.}
\label{fig:testRavi}
\end{figure*}

In this section, we evaluate the individual performance of our angular SR network. For this experiment, the 
angular resolution of the test images are downsampled from 14x14 to 7x7 while keeping the spatial resolution at 374x540 pixels. These low-angular images are then input to the angular SR network to bring them back to the original angular resolution. The network is trained for each color channel separately. We compare our method against Kalantari \textit{et al.} \cite{LearningViewSynthesis}, which is a very recent convolutional neural network based novel view synthesis method, and against Wanner and Goldluecke \cite{wanner2014variational}, which utilizes disparity maps in a variational optimization framework. Wanner and Goldluecke \cite{wanner2014variational} may work with any disparity map generation algorithm; thus, we report results with the disparity generation algorithms given in  \cite{wanner2012globally},  \cite{Tao_2013_ICCV}, \cite{wang2015occlusion}, and \cite{jeon2015accurate}.  In Table \ref{tab:table2}, we quantitatively compared the results with the state-of-the-art angular resolution enhancement methods using PSNR and SSIM. In Figure \ref{fig:testRavi}, we provide a visual comparison. The scene contains occluded regions, which are generally difficult for view synthesis. Our angular SR method produces significantly better results compared to all other approaches. 

Finally, we would like to note that the angular SR network, by itself, may turn out to be useful, since it may be combined with any single-image resolution enhancement method to enhance the spatial and angular resolution of a light field capture.

\begin{table*}
\caption{Comparison of different methods for angular resolution enhancement.}
  \centering  
  \begin{tabular}{|c|c|cccc|c|c|}
    \toprule
     &  &  \multicolumn{4}{c|}{Wanner and Goldluecke \cite{wanner2014variational}} & & \\
   	Picture name & Evaluation metric & \multirow{2}{5em}{\centering Disparity \cite{wanner2012globally}}& \multirow{2}{5em}{\centering Disparity \cite{Tao_2013_ICCV}}& \multirow{2}{5em}{\centering Disparity \cite{wang2015occlusion}} & \multirow{2}{5em}{\centering Disparity \cite{jeon2015accurate}} & \multirow{2}{5em}{\centering Kalantari \textit{et al.} \cite{LearningViewSynthesis}} & \multirow{2}{5em}{\centering Angular SR network} \\
   	& & & & & & &  \\
    \midrule
    \multirow{2}{5em}{\textit{Flower 1}} & PSNR (dB) & 22.03& 29.52& 24.39& 28.21& 33.31& \redbf{35.95}\\
     & SSIM & 0.789& 0.941& 0.910& 0.934& 0.969& \redbf{0.982}\\ \hline 
    \multirow{2}{5em}{\textit{Cars}} & PSNR (dB) & 19.74& 27.27& 22.09& 27.51& 31.65& \redbf{35.21}\\
     & SSIM & 0.792& 0.946& 0.911& 0.949& 0.966& \redbf{0.983}\\ \hline
     \multirow{2}{5em}{\textit{Flower 2}} & PSNR (dB) & 20.61& 27.56& 23.65& 27.04& 31.93& \redbf{36.75}\\
     & SSIM & 0.645& 0.919& 0.899& 0.924& 0.959& \redbf{0.980}\\ \hline
     \multirow{2}{5em}{\textit{Rock}} & PSNR (dB) & 16.57& 30.46& 30.55& 30.21& \redbf{34.67}& 34.09\\
     & SSIM & 0.488& 0.945& 0.948& 0.946& \redbf{0.970}& 0.963\\ \hline
     \multirow{2}{5em}{\textit{Leaves}} & PSNR (dB) & 15.03& 23.54& 20.08& 23.88& 27.80& \redbf{33.08}\\
     & SSIM & 0.481& 0.882& 0.855& 0.893& \redbf{0.963}& 0.956\\ 
    \bottomrule
  \end{tabular}
  \label{tab:table2}
\end{table*}

\subsection{Depth Map Estimation Accuracy}

\begin{figure*}
  \subfloat[\label{fig:DepthOR1}]
  {\includegraphics[width=.35\textwidth]{GT11}}\hfill
  \subfloat[\label{fig:Depth7}]
  {\includegraphics[width=.31\textwidth]{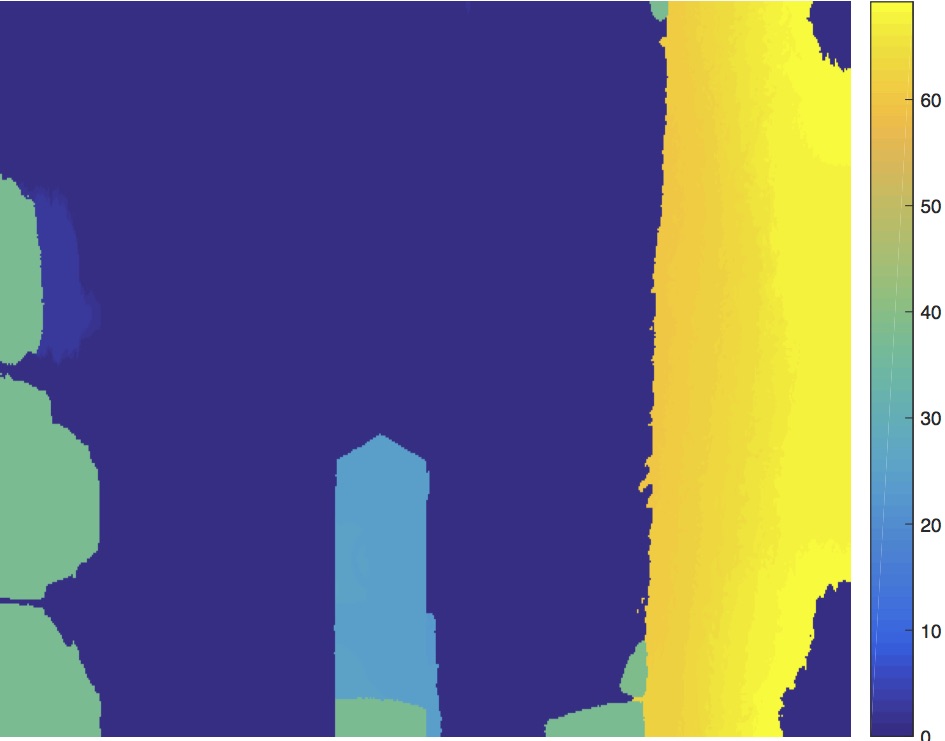}}\hfill
  \subfloat[\label{fig:Depth14}]
  {\includegraphics[width=.31\textwidth]{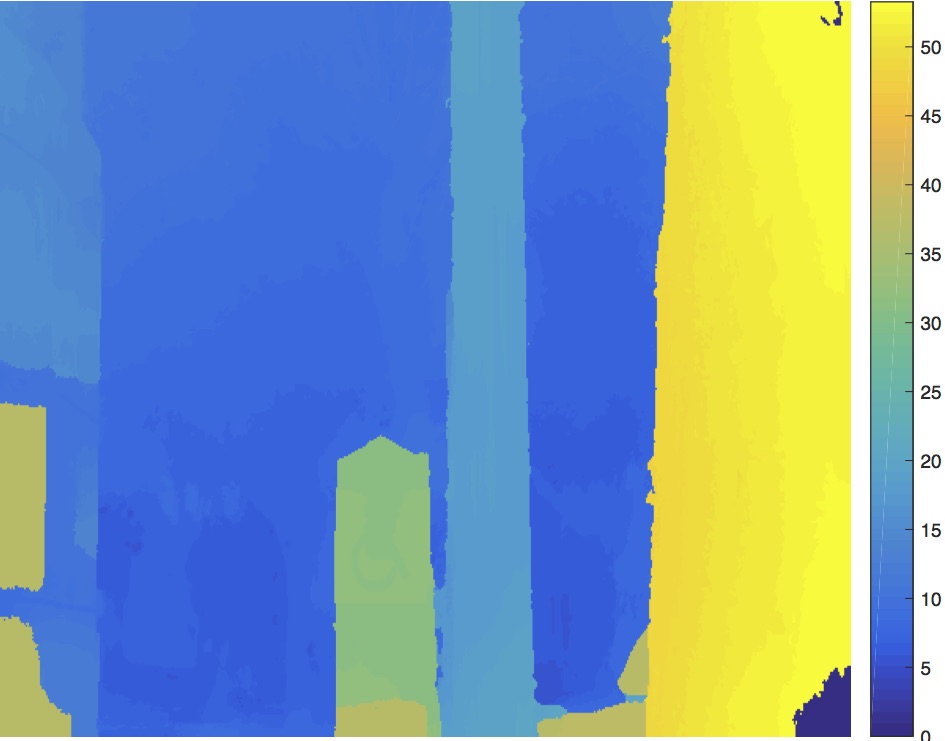}}\hfill

\caption{Depth map estimation accuracy. (a) Middle perspective image. (b) Estimated depth map from the input light field with 7x7 angular resolution. (c) Estimated depth map from enhanced light field with 14x14 angular resolution.}
\label{fig:Depth}
\end{figure*}

\begin{figure*}
  \subfloat[\label{fig:DepthOR}]
  {\includegraphics[width=.35\textwidth]{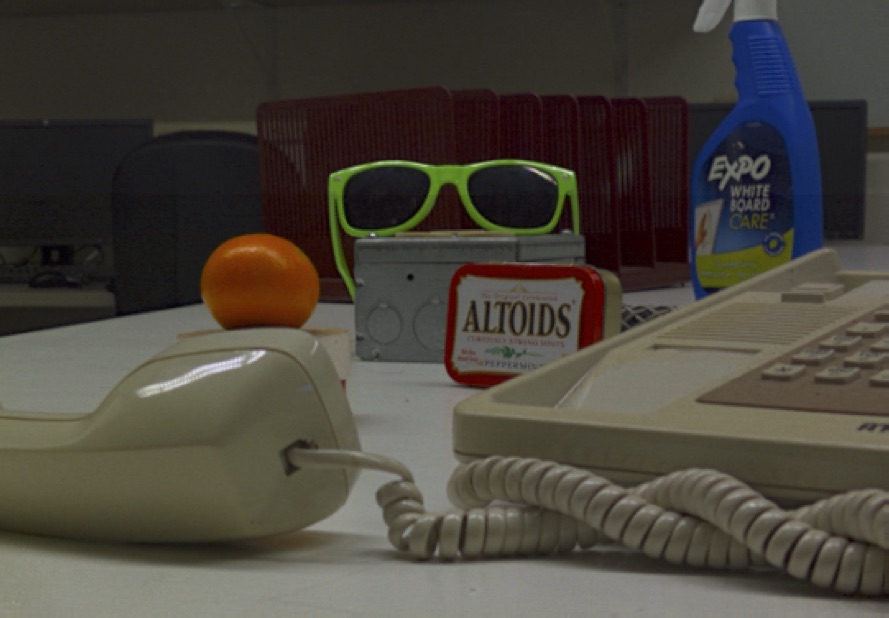}}\hfill
  \subfloat[\label{fig:Depth7_1}]
  {\includegraphics[width=.31\textwidth]{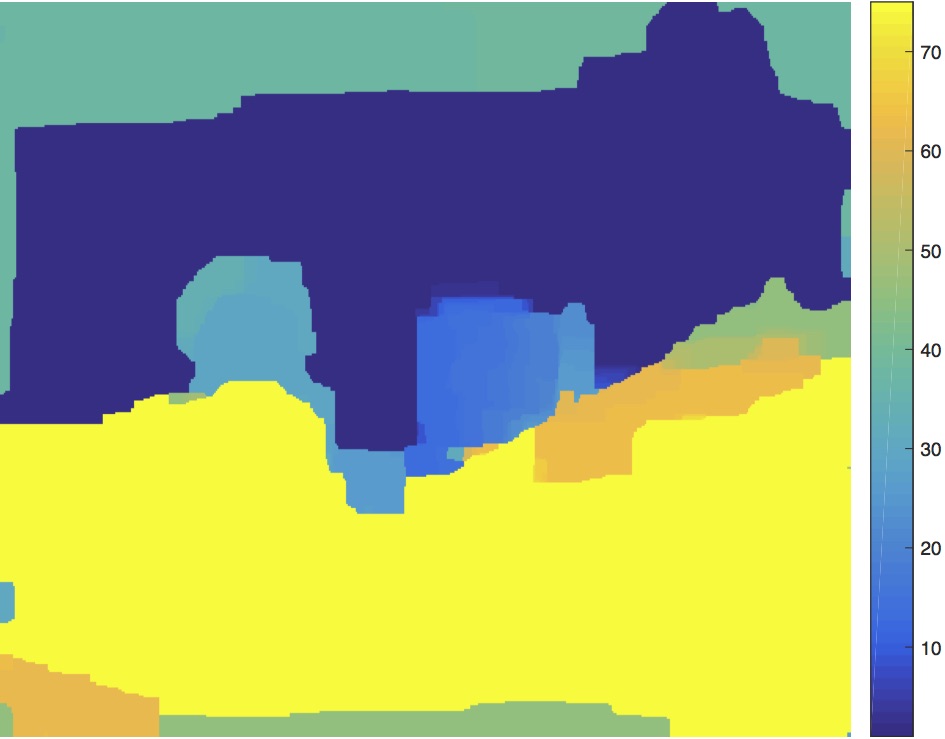}}\hfill
  \subfloat[\label{fig:Depth14_1}]
  {\includegraphics[width=.31\textwidth]{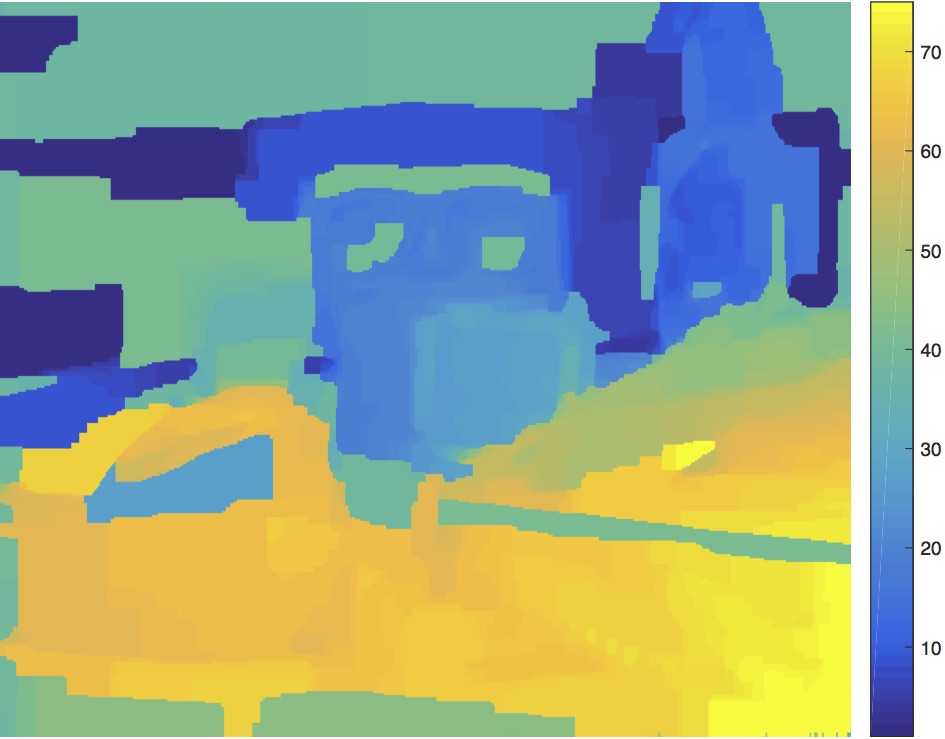}}\hfill
\caption{Depth map estimation accuracy. (a) Middle perspective image. (b) Estimated depth map from the input light field with 7x7 angular resolution. (c) Estimated depth map from enhanced light field with 14x14 angular resolution.}
\label{fig:Depth_1}
\end{figure*}

One of the capabilities of light field imaging is depth map estimation, whose accuracy is directly related to the angular resolution of light field. In Figure \ref{fig:Depth} and Figure \ref{fig:Depth_1}, we compare depth maps obtained from the input light fields and the light fields enhanced by the proposed method. The depth maps are estimated using the method in \cite{jeon2015accurate}, which is specifically designed for light fields. It is clearly seen that depth maps obtained from light fields enhanced with the proposed method show higher accuracy. With the enhanced light fields, even close depths can be differentiated, unlike the low-resolution light fields. 

\subsection{Model and Performance Trade-Offs}
To evaluate the trade-off between performance and speed, and to investigate the relation between performance and the network parameters, we modify different parameters of the network architecture and compare with the base architecture. All the experiments are performed on a machine with Intel Xeon CPU E5-1650 v3 3.5GHz, 16GB RAM and Nvidia 980ti 6GB graphics card.  

\subsubsection*{Filter Size}
\begin{figure}
   \includegraphics[width=0.48\textwidth]{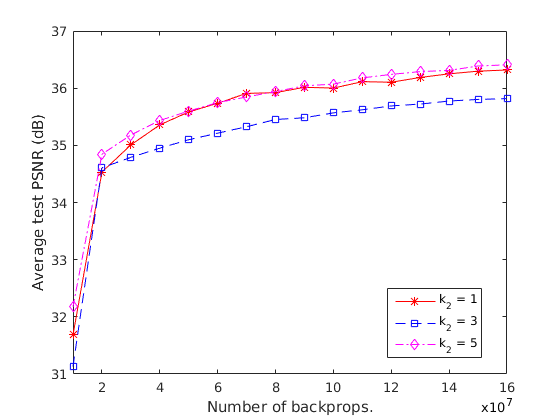}
      \caption{Effect of the filter size on performance.}
      \label{filtersize}
\end{figure}

In the proposed spatial SR network, the filter sizes in the two convolution layers are $k_1$ = 3 and $k_2$ = 1, respectively. The filter size of the first convolution layer is kept at $k_1$ = 3; this means, for each light ray (equivalently, perspective image), the network is considering the light rays (perspective images) in a 3x3 neighborhood in the first convolution layer. Since higher dimensional relations are taken care of in the second convolution layer, and since keeping the filter size small minimizes the boundary effects---note that the input size in the first layer is 14x14---this seems to be a reasonable choice for the first layer. On the other hand, we have more flexibility in the second convolution layer. We examined the effect of the filter size in the second convolution layer by setting $k_2$ = 3 and $k_2$ = 5 while keeping the other parameters intact. In Figure \ref{filtersize}, we provide the average PSNR values on the test dataset for different values of $k_2$ as a function of training backpropagation numbers. When $k_2 =5$, the convergence is slightly better than the case with $k_2 =1$. In Table \ref{table:timeSR}, we show the final PSNR values and the computation times per channel (namely, the $red$ channel) for a perspective image. It is seen that while the PSNR is slightly improved the computation time is more than doubled when we increase the filter size from $k_2=1$ to $k_2=5$.  

\begin{table}
\caption{Effect of the filter size on performance and the speed of the spatial SR network.}
  \centering 
  \begin{tabular}{c|c|c}
    \toprule
     Filter size & PSNR (dB) & Time (sec) \\
    \midrule
     $k_2$ = 1 & 36.31 & 17.58 \\
     $k_2$ = 3 & 35.81 & 27.62 \\
     $k_2$ = 5 & 36.40 & 45.18 \\ \hline
  \end{tabular}
   \label{table:timeSR}
\end{table} 
\vspace{2mm}
\subsubsection*{Number of Layers and Number of Filters}
\begin{figure}
   \includegraphics[width=0.48\textwidth]{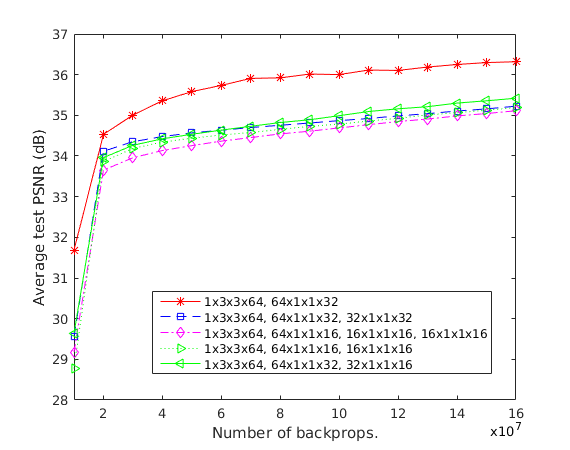}
      \caption{Effect of number of layers and number of filters on performance.}
      \label{layerandfilter}
\end{figure}

We also examine the network performance for different number of layers and different number of filters. We implemented deeper architectures by adding new convolution layers after the second convolution layer. The three-layer network presented in the previous section is compared against the four-layer and five-layer networks. For the four-layer network, we evaluated the performance for different filter combinations. The network configurations we used are shown in Table \ref{table:NLFN}. In Figure \ref{layerandfilter}, we provide the convergence curves for these different network configurations. We observe that the simple three-layer network performs better than the others. This means that increasing the number of convolution layers is causing overfitting and degrading the performance. 

\begin{table}
\caption{Different network configurations used to evaluate the performance of the spatial SR network.}  
  \centering 
  \begin{tabular}{c|c|c|c|c}
  \hline
       & \multicolumn{4}{c}{Convolution Layers} \\ \cline{2-5}
    & \multirow{2}{5em}{\centering First} & \multirow{2}{5em}{\centering Second} & \multirow{2}{5em}{\centering Third} & \multirow{2}{5em}{\centering Fourth}\\ 
    & & & & \\ \hline
    3 layer & 1x3x3x64 & 64x1x1x32 & - & - \\
    4 layer & 1x3x3x64 & 64x1x1x32 & 32x1x1x32 & - \\
    4 layer & 1x3x3x64 & 64x1x1x16 & 16x1x1x16 & - \\
    4 layer & 1x3x3x64 & 64x1x1x32 & 32x1x1x16 & - \\
    5 layer & 1x3x3x64 & 64x1x1x16 & 16x1x1x16 & 16x1x1x16 \\ \hline
  \end{tabular}
  \label{table:NLFN}
\end{table}

\subsection{Further Increasing the Spatial Resolution}

\begin{figure*}
\centering
\captionsetup{justification=centering,margin=0cm}
  {\includegraphics[width=.19\textwidth]{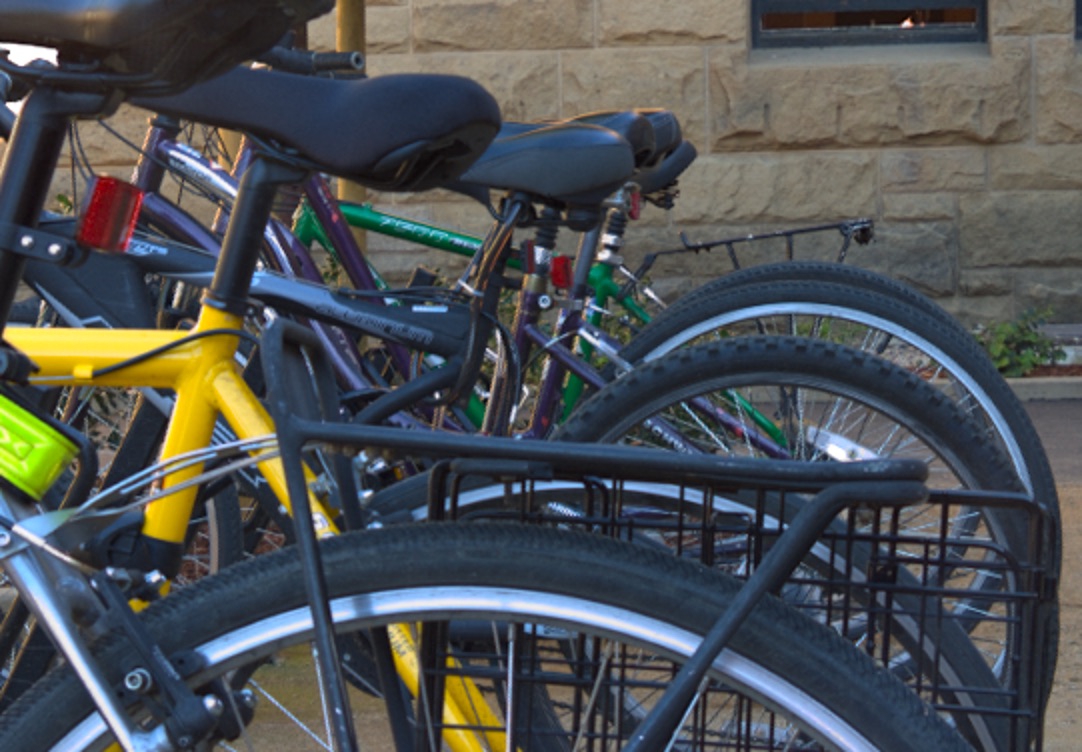}}\hfill
  {\includegraphics[width=.19\textwidth]{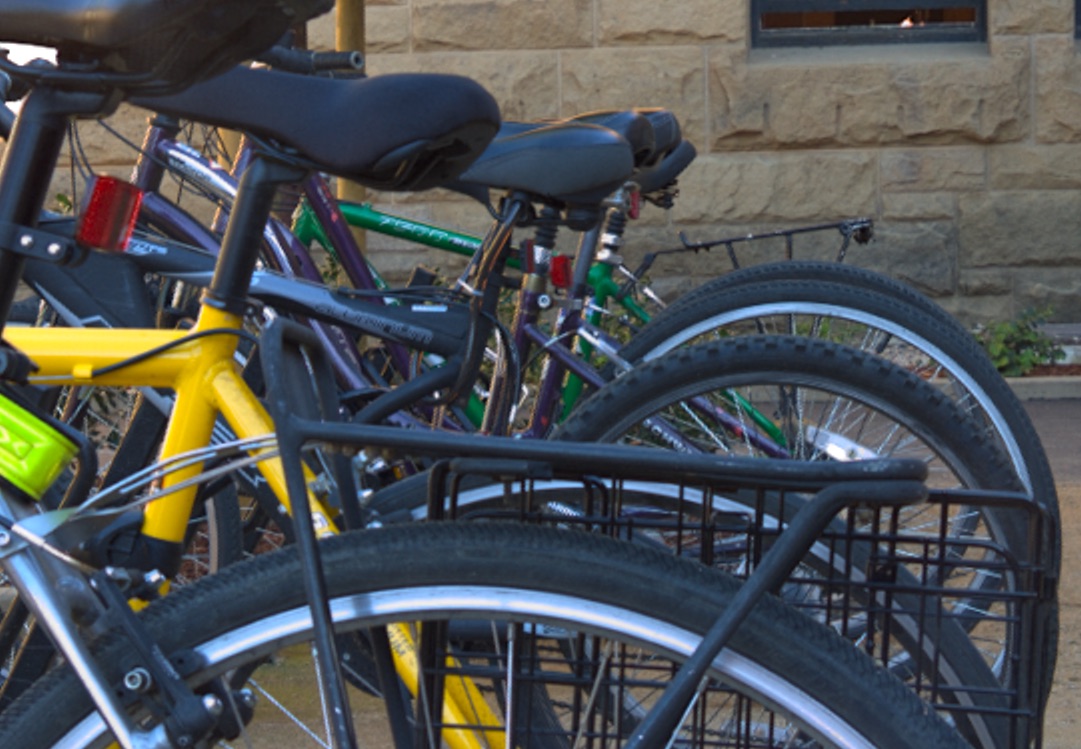}}\hfill
  {\includegraphics[width=.19\textwidth]{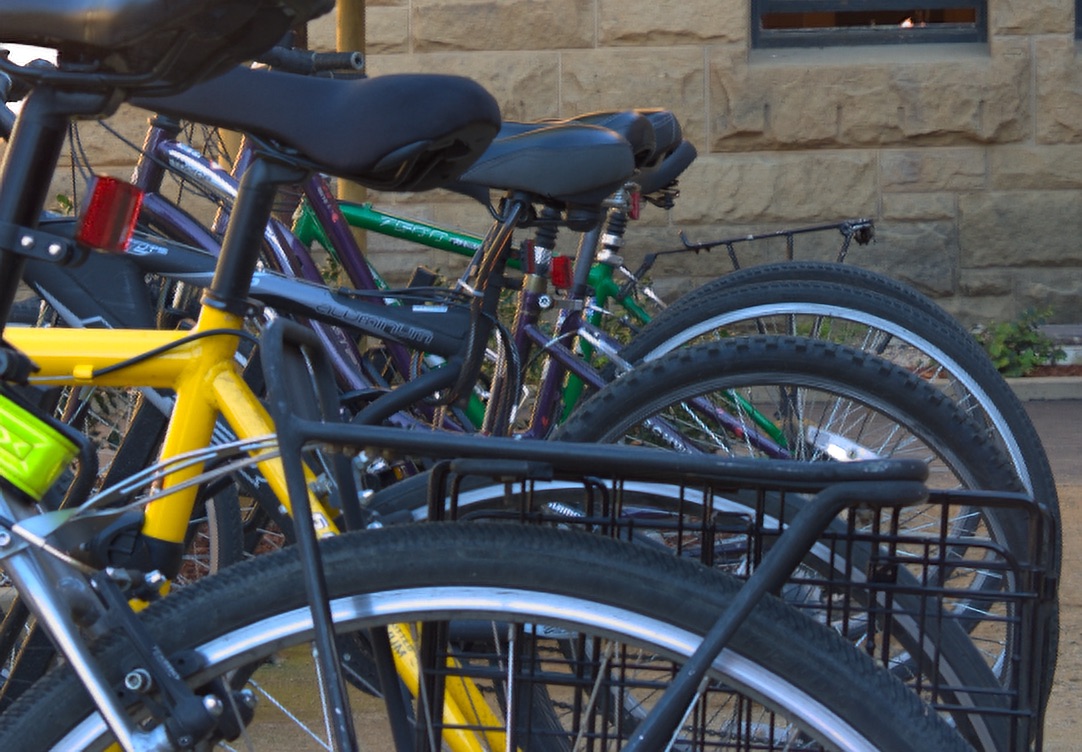}}\hfill
  {\includegraphics[width=.19\textwidth]{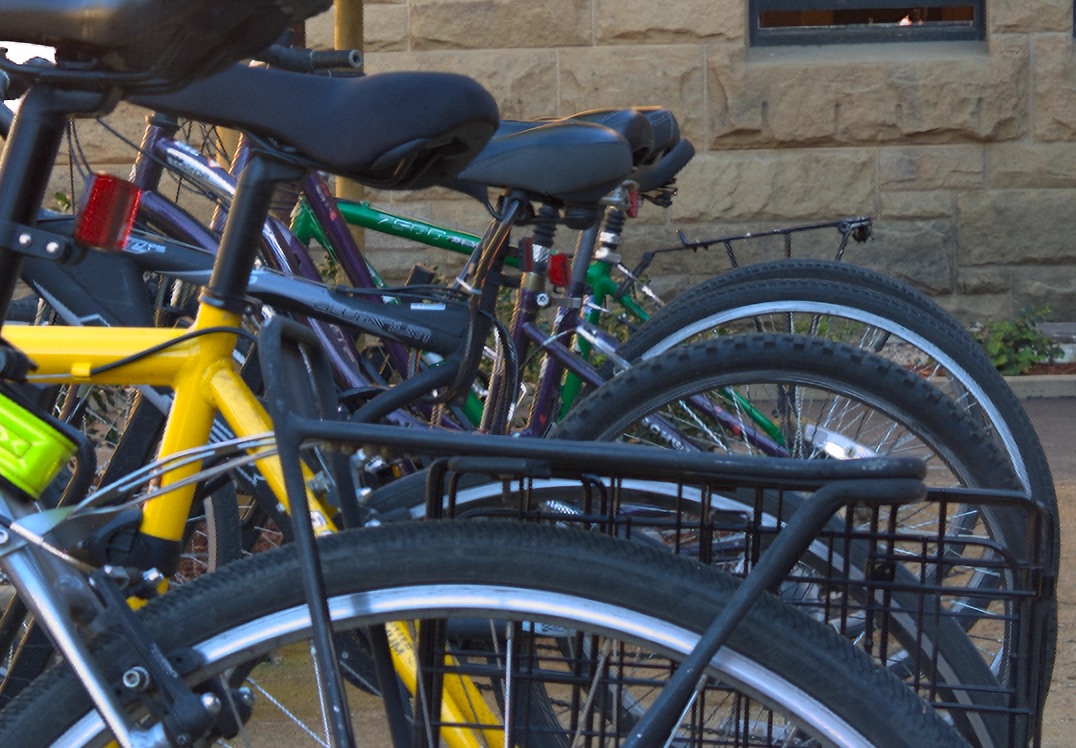}}\hfill
  {\includegraphics[width=.19\textwidth]{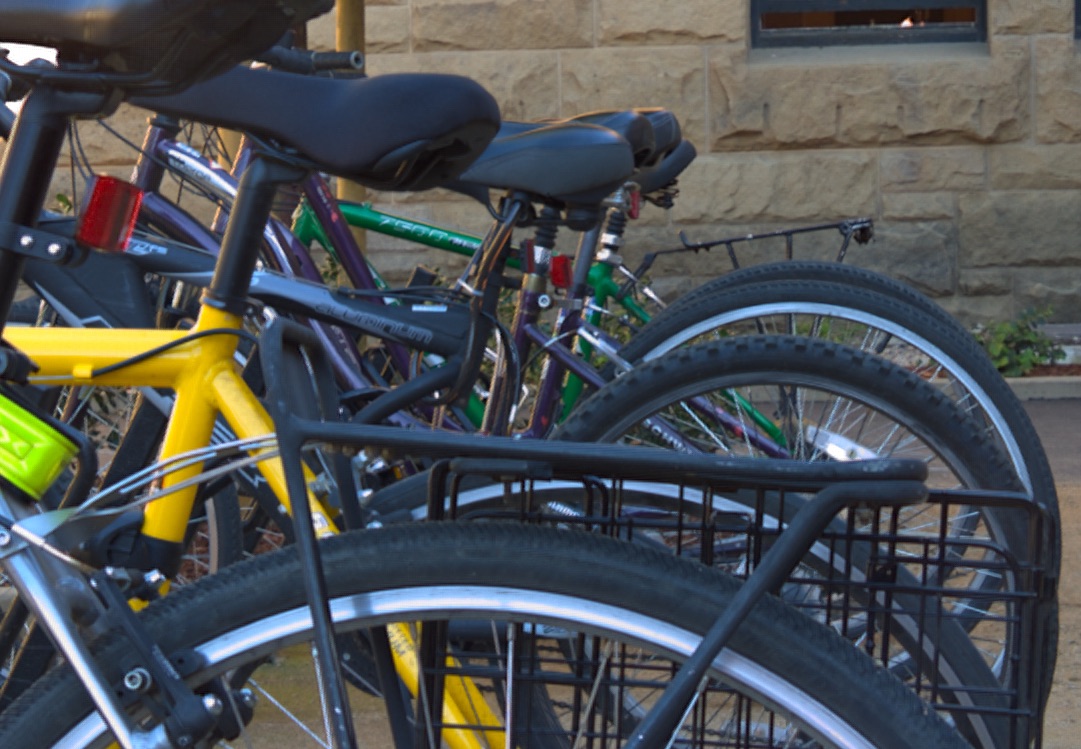}}
  
  \subfloat[Bicubic resizing (\textit{imresize}) \label{fig:test61}]
  {\includegraphics[trim= 705 290 276 359, clip=true, width=0.19\textwidth]{Bicubic_new2}}\hfill
  \subfloat[Bicubic interpolation \label{fig:test62}]
  {\includegraphics[trim= 705 288 276 359, clip=true, width=0.19\textwidth]{B2}}\hfill
  \subfloat[LFCNN \cite{yoon2015learning} \label{fig:test63}]
  {\includegraphics[trim= 705 290 276 359, clip=true, width=0.19\textwidth]{LFCNN2}}\hfill
  \subfloat[DRRN \cite{Tai-DRRN-2017} \label{fig:test64}]
  {\includegraphics[trim= 705 292 276 359, clip=true, width=0.19\textwidth]{10_x2}}\hfill
  \subfloat[Proposed LFSR \label{fig:test65}]
  {\includegraphics[trim= 705 288 276 359, clip=true, width=0.19\textwidth]{P2}}

  {\includegraphics[width=.19\textwidth]{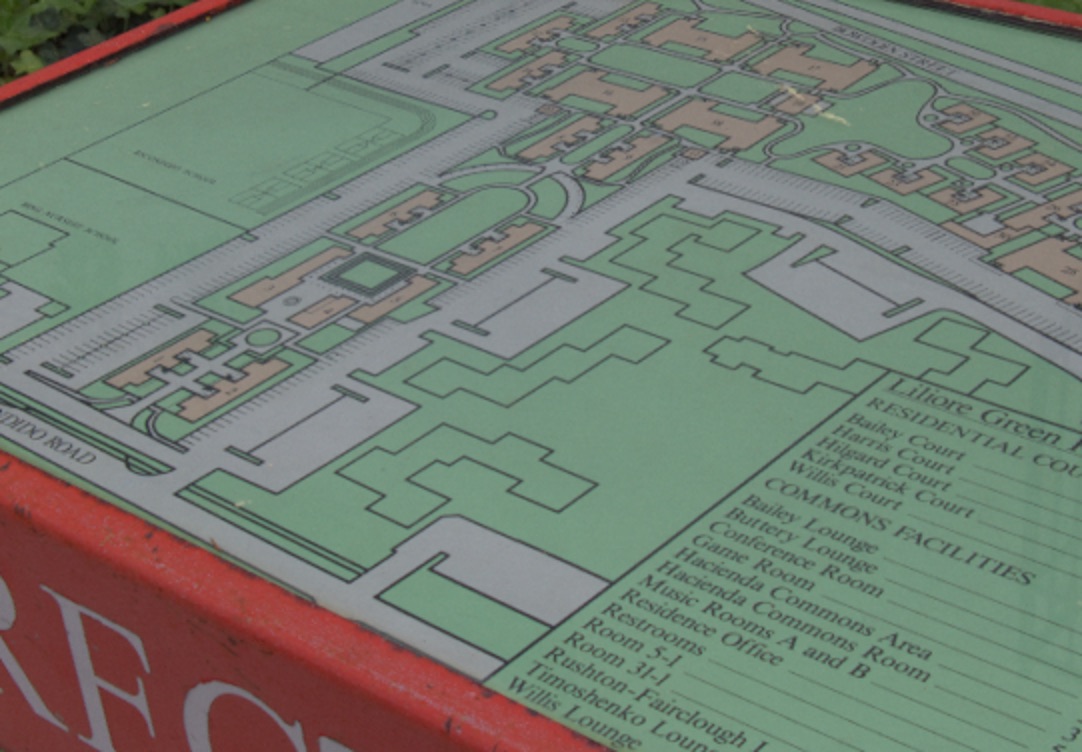}}\hfill
  {\includegraphics[width=.19\textwidth]{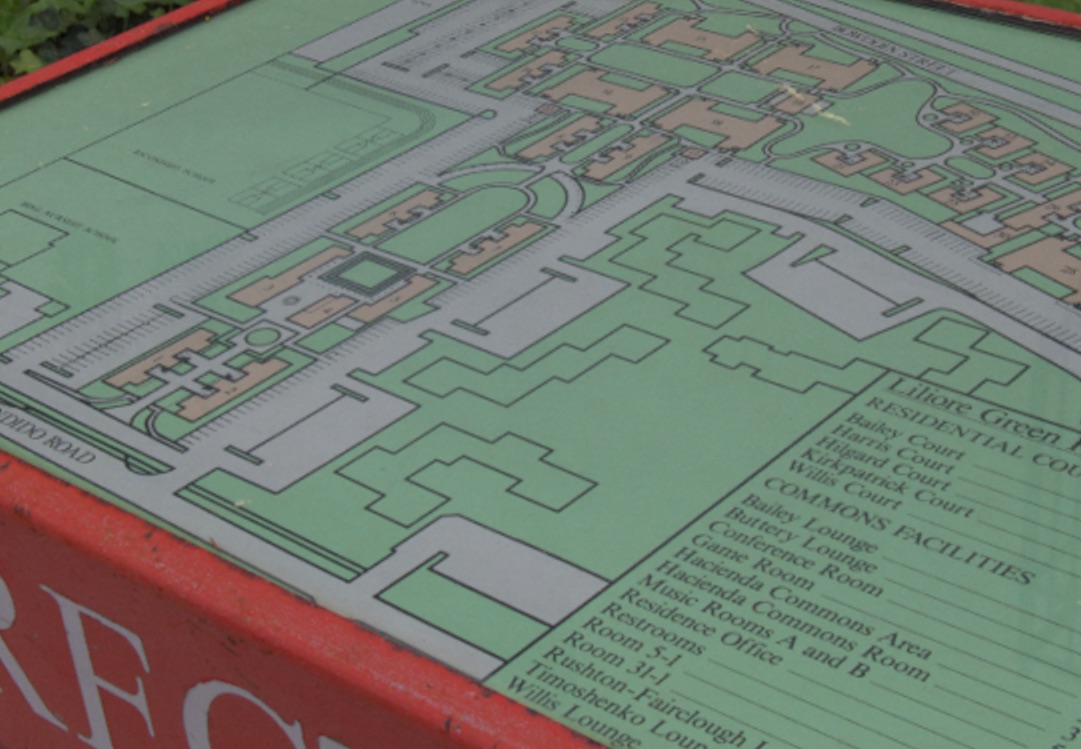}}\hfill
  {\includegraphics[width=.19\textwidth]{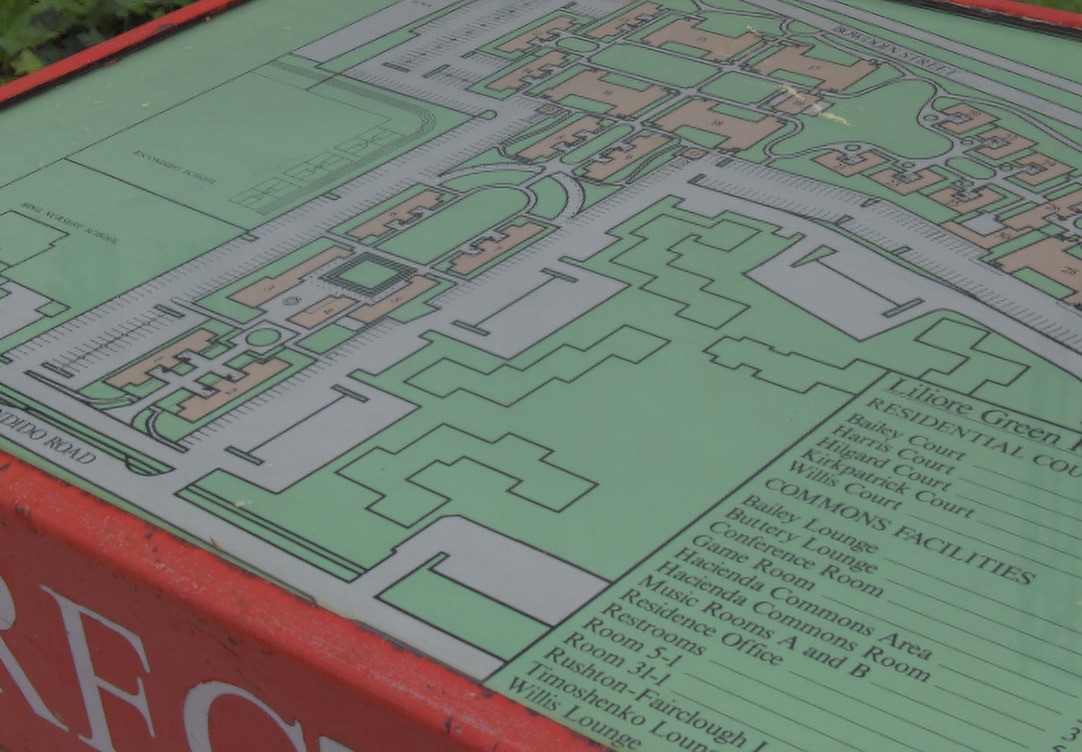}}\hfill
  {\includegraphics[width=.19\textwidth]{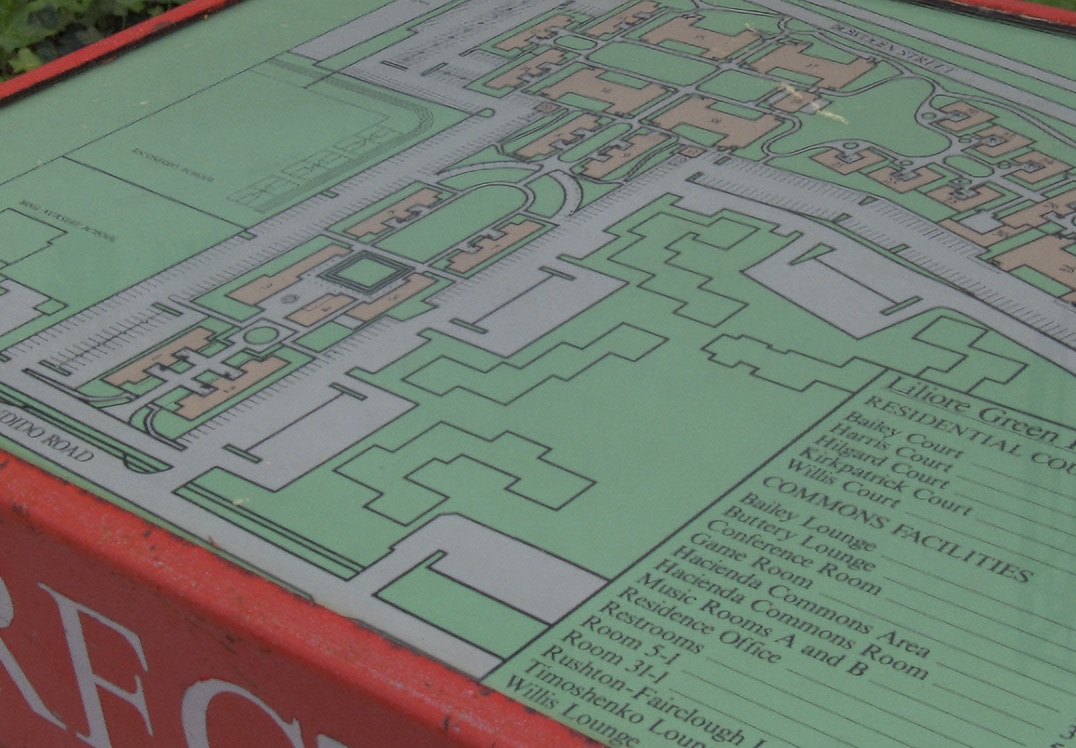}}\hfill
  {\includegraphics[width=.19\textwidth]{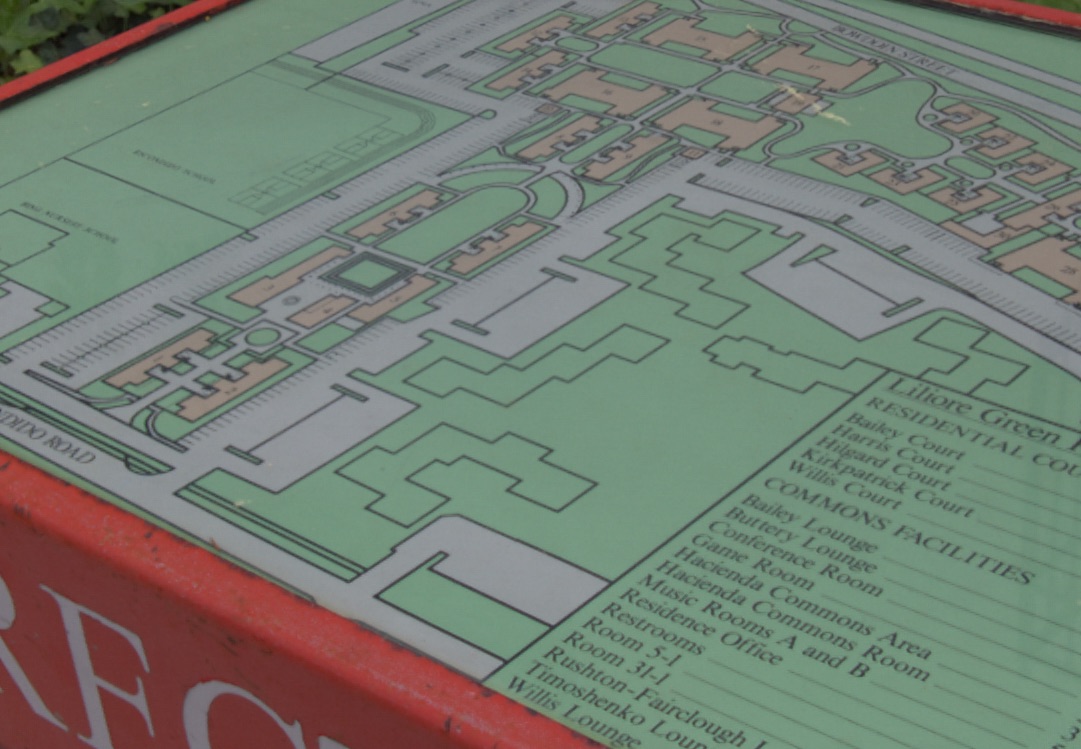}}

  \vspace{1em}
  {\includegraphics[trim= 410 645 600 10, clip=true, width=0.19\textwidth]{Bicubic_new19}}\hfill
  {\includegraphics[trim= 410 643 600 10, clip=true, width=0.19\textwidth]{B19}}\hfill
  {\includegraphics[trim= 410 645 600 10, clip=true, width=0.19\textwidth]{LFCNN19}}\hfill
  {\includegraphics[trim= 411 647 600 12, clip=true, width=0.19\textwidth]{35_x2}}\hfill
  {\includegraphics[trim= 410 643 600 10, clip=true, width=0.19\textwidth]{P19}}\hfill    
  \subfloat[Bicubic resizing (\textit{imresize}) \label{fig:test68}]
  {\includegraphics[trim= 900 590 90 50, clip=true, width=0.19\textwidth]{Bicubic_new19}}\hfill
  \subfloat[Bicubic interpolation \label{fig:test69}]
  {\includegraphics[trim= 900 588 90 50, clip=true, width=0.19\textwidth]{B19}}\hfill
  \subfloat[LFCNN \cite{yoon2015learning}\label{fig:test70}]
  {\includegraphics[trim= 900 590 90 50, clip=true, width=0.19\textwidth]{LFCNN19}}\hfill
  \subfloat[DRRN \cite{Tai-DRRN-2017}\label{fig:test71}]
  {\includegraphics[trim= 900 593 90 50, clip=true, width=0.19\textwidth]{35_x2}}\hfill
  \subfloat[Proposed LFSR \label{fig:test72}]
  {\includegraphics[trim= 900 588 90 50, clip=true, width=0.19\textwidth]{P19}}\hfill  
\caption{Visual comparison of different methods.}
\label{fig:test4}
\end{figure*}

For quantitative evaluation, we need to have the ground truth; thus, we downsample the captured light field to generate its lower resolution version. In addition, we can visually evaluate the performance of the proposed method without downsampling and further increasing the spatial resolution of the original images. In Figure \ref{fig:test4}, we provide a comparison of bicubic resizing, bicubic interpolation, the LFCNN method \cite{yoon2015learning}, the DRRN method \cite{Tai-DRRN-2017}, and the proposed LFSR method. The spatial resolution of each perspective image is increased from 374x540 to 748x1080. The results of the proposed method seem to be preferable over the others with less artifacts. The LFCNN results in sharp images but has some visible artifacts. The DRNN method seems to distort some texture, especially visible in the second example image, while the proposed method preserves the texture well. 

\section{Discussion and Conclusions}

In this paper, we presented a convolutional neural network based light field super-resolution method. The method consists of two separate convolutional neural networks trained through supervised learning. The architecture of these networks are composed of only three layers, reducing computational complexity. The proposed method shows significant improvement both quantitatively and visually over the baseline bicubic interpolation and another deep learning based light field super-resolution method. In addition, we compared the angular resolution enhancement part of our method against two methods for novel view synthesis. We also demonstrated that enhanced light field results in more accurate depth map estimation due to the increase in angular resolution. 

The spatial super-resolution network is designed to generate one perspective image. One may suggest to generate all perspectives in a single run; however, this would result in a larger network, requiring larger size dataset and more training. Instead, we preferred to have a simple, specialized, and effective architecture.  

Similar to other neural network based super-resolution techniques, the method is designed to increase the resolution by an integer factor (two). It can be applied multiple times to increase the resolution by factors of two. A non-integer factor size change is also possible by first interpolating using the proposed method and then downsampling using a standard technique.

The network parameters are optimized for a specific light field camera. For different cameras, the specific network parameters, such as filter dimensions, may need to be optimized. We, however, believe that the overall architecture is generic and would work well with any light field imaging system once optimized.

\section*{Acknowledgements}

This work is supported by TUBITAK Grant 114E095.

\ifCLASSOPTIONcaptionsoff
  \newpage
\fi



%

\end{document}